\documentclass[lettersize,journal]{IEEEtran}

\usepackage{amsmath,amsfonts}
\usepackage{algorithmic}
\usepackage{array}
\usepackage[caption=false,font=normalsize,labelfont=sf,textfont=sf]{subfig}
\usepackage{textcomp}
\usepackage{stfloats}
\usepackage{url}
\usepackage{verbatim}
\usepackage{balance}

\usepackage{times} 
\usepackage{amssymb} 
\usepackage{tabularx} 
\usepackage{multirow}
\usepackage{booktabs}
\usepackage{scalerel}

\usepackage{enumitem} 
\usepackage{float}
\usepackage{dsfont} 
\usepackage{colortbl} 
\usepackage{multicol} 
\usepackage[table]{xcolor}

\usepackage{caption}
\usepackage{graphicx} 
\usepackage{hyperref} 

\definecolor{lightgray}{gray}{0.9}

\newcommand{\ie}{{i.e.}}
\newcommand{\eg}{{e.g.}}
\newcommand{\etc}{{etc}}

\graphicspath{{./}}
\DeclareGraphicsExtensions{.pdf, .jpg, .png}
\newcommand{\Fig}[1]{Figure \ref{fig:#1}}

\newcommand{\bbest}[1]{\bf{#1}}
\newcommand{\gbbest}[1]{\bbest{\underline{#1}}}

\newcommand{\stdx}[1]{\scaleto{\textcolor{gray}{\pm #1}}{1.3ex}}
\newcommand{\UDAbackbone}[1]{\scaleto{\textcolor{gray}{#1}}{1.3ex}}

\newcommand{\thedataset}{UrbanSyn}

\usepackage{lineno}

\begin{document}


\title{All for One, and One for All: UrbanSyn Dataset, the third Musketeer of Synthetic Driving Scenes}

\author{
Jose L. G\'{o}mez$^{1 \ast}$, Manuel Silva$^{1,5\ast}$, Antonio Seoane $^{5}$, Agnès Borr\'{a}s $^{1}$, \\ 
Mario Noriega $^{1}$, German Ros $^{3}$, Jose A. Iglesias-Guitian $^{2,5}$, Antonio M. López $^{1,4}$\\
{\small
$^{1}$ Computer Vision Center (CVC), $^{2}$ CITIC - Centre for ICT Research, $^{3}$ NVIDIA, \\ $^{4}$ Computer Science Department, Universitat Autònoma de Barcelona (UAB),\\
$^{5}$ Universidade da Coruña (UDC)\\

Contact emails: \texttt{jlgomez@cvc.uab.cat, m.silva1@udc.es, antonio.seoane@udc.es, agnes.borras@cvc.uab.cat, mnoriega@cvc.uab.cat, grossanchez@nvidia.com, j.iglesias.guitian@udc.es, antonio@cvc.uab.es} \\
}
\thanks{
\textbf{Funding Acknowledgements:} This work has been supported by the Spanish grants Ref. PID2020-115734RB-C21 (ADA/SSL-ADA subproject) and PID2020-115734RB-C22 (ADA/PGAS-ADA subproject) both funded by MCIN/AEI/10.13039/501100011033. 
Antonio M. López acknowledges the financial support to his general research activities given by ICREA under the ICREA Academia Program. CVC's authors acknowledge the support of the Generalitat de Catalunya CERCA Program and its ACCIO agency to CVC’s general activities.
Jose A. Iglesias-Guitian acknowledges the financial support to his general research activities given by UDC-Inditex InTalent programme, the Spanish Ministry of Science and Innovation (AEI/RYC2018-025385-I), and Xunta de Galicia (ED431F 2021/11, EU-FEDER ED431G 2019/01).}
\thanks{$^\ast$ Jose L. G\'{o}mez and Manuel Silva are co-first authors.}
}


\makeatletter
\g@addto@macro\@maketitle{
    \setcounter{figure}{0} 
	\begin{figure}[H]
		\setlength{\linewidth}{\textwidth}
		\setlength{\hsize}{\textwidth}
		\vspace{-5mm}
		\centering
		\includegraphics[width=\textwidth]{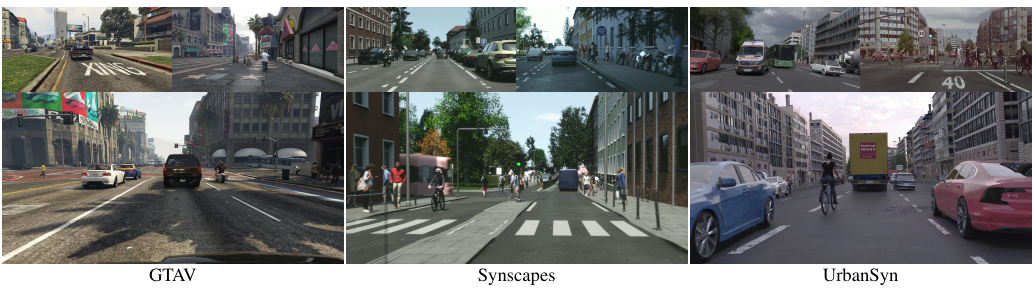}
        \caption{Samples from the Musketeers dataset: GTAV \cite{richter2016playing}, Synscapes \cite{wrenninge2018synscapes}, and {\thedataset} (ours).}        
        \label{fig:short}
	\end{figure}
}
\makeatother

\maketitle


\begin{abstract}
We introduce UrbanSyn, a photorealistic dataset acquired through semi-procedurally generated synthetic urban driving scenarios. Developed using high-quality geometry and materials, UrbanSyn provides pixel-level ground truth, including depth, semantic segmentation, and instance segmentation with object bounding boxes and occlusion degree. 
It complements GTAV and Synscapes datasets to form what we coin as the 'Three Musketeers'. We demonstrate the value of the Three Musketeers in unsupervised domain adaptation for image semantic segmentation. Results on real-world datasets, Cityscapes, Mapillary Vistas, and BDD100K, establish new benchmarks, largely attributed to UrbanSyn. We make UrbanSyn openly and freely accessible.
\end{abstract}

\section{Introduction}

Human data labeling is a major bottleneck for training and validating many deep perception models. To solve vision tasks, datasets of synthetic images have attracted a great deal of interest \cite{gaidon2016vkitti, ros2016synthia, richter2016playing, wrenninge2018synscapes}. They come with many types of automatically generated ground truths ({\ie}, labels), and, in contrast to human-produced labels, synthetic ones are accurate and consistent across images. However, since deep models trained on synthetic images must later perform in real-world images, the synth-to-real domain gap must be addressed, which remains an open research topic \cite{wang2018survey, wilson2020survey, csurka2022visual}. 
In fact, domain gaps also exist among real-world images captured by different camera sensors. However, synth-to-real domain adaptation is especially relevant because it opens the door to minimizing costly and not fully safe autonomous driving developments in the real world. For instance, in an autonomous driving simulator, we can force desired traffic scenarios that are difficult to observe in the real world by random roaming and train perception models with data captured during such simulations. Research on synth-to-real domain adaptation pursues to develop methods to adapt these models to perform in the real world, preferably in an unsupervised manner.
As argued in other AI sub-fields \cite{yang2019neuralhype,ferrari2019nworrying}, we must rely on reasonably strong baselines to avoid illusory gains when developing new proposals for a given task. In our case, this means starting with reasonably accurate synthetic-only baseline models to bridge the synth-to-real domain gap. The prevailing approach to addressing the synth-to-real gap typically involves focusing on a single synthetic dataset at a time, which leads to less-than-optimal results. It is important to note that when dealing exclusively with real-world labeled images, training complex models on datasets compiled from various sources has been shown to enhance their accuracy compared to training them on each source separately. This increase in accuracy is not produced by the larger number of training samples but by the additional diversity derived from the use of different data sources. This has been observed for core visual tasks such as semantic segmentation \cite{lambert2020Mseg}, object detection \cite{zhou2022multidatasetdetect}, and depth estimation \cite{ranftl2022MDEmixingdatasets}. 

Driven by the idea of multi-source approaches \cite{gong2021mDALU, zhao2019multisource, he2021multisource,gomez2023cotraining}, we introduce the {\thedataset} dataset. A collection of highly realistic synthetic images of urban scenarios generated semi-procedurally. {\thedataset} targets computer vision tasks in driving domains and features labels, including pixel-wise instance and semantic masks, depth, and bounding boxes for objects with occlusion degree. This novel dataset aims at complementing other publicly available synthetic datasets such as GTAV \cite{richter2016playing} and Synscapes \cite{wrenninge2018synscapes}, addressing the synth-to-real domain gap from a multi-source perspective. 
To achieve such a complementarity, {\thedataset} is based on unbiased path tracing instead of the video game deferred rendering technique utilized in GTAV. This results in rendering real-world imperfections in a more realistic manner. Moreover, while Synscapes is also more realistic than GTAV, {\thedataset} generation goes beyond by simulating atmospheric effects through physics-based light-transport simulation in participating media. Additionally, {\thedataset} is generated from totally new 3D environments and assets. All this means that {\thedataset} images show scenes different from those in GTAV and Synscapes in terms of content and appearance.

In this paper, we demonstrate the value of {\thedataset} by addressing semantic segmentation, a challenging task~\cite{garcia2018survey, csurka2022semsegreview, Mo2022surveysemseg} that has proven to be key for the correct performance of many autonomous driving solutions~\cite{janai2020cvadsurvey, tampuu2020etesurvey, li2022survey}. 
Note that onboard semantic segmentation involves assigning a traffic-relevant semantic class ({\eg}, Vehicle, Pedestrian, Road, etc.) to each pixel of an image captured for the automatic understanding of the vehicle's surroundings and so for performing safe driving maneuvers. Deep learning models are the most effective way to accomplish this visual task. Thus, it is essential to gather training images that bring diversity and accurately labeled semantic information. However, obtaining this labeled data is a time-consuming process, taking 60-90 minutes per image due to the manual human labeling involved. {\thedataset} brings accurate pixel-wise semantic labels for onboard images depicting diverse urban environments and objects.
%
%
When used individually, we show that {\thedataset} gives rise to stronger semantic segmentation baselines than GTAV and Synscapes datasets. But most importantly, the strongest baselines arise when GTAV, Synscapes, and {\thedataset} are combined. Inspired by Alexandre Dumas, we refer to the combination of these datasets as \emph{The Three Musketeers} \footnote{All synthetic datasets for one, and one for all.} (\Fig{short}). Our experiments make use of modern and relevant approaches, including DeepLab V3+~\cite{chen2017DeepLab} and SegFormer~\cite{xie2021segformer}, evaluated in challenging real-world datasets, such as Cityscapes \cite{cordts2016cityscapes}, BDD100K \cite{yu2020BDD}, and Mapillary Vistas \cite{neuhold2017mapillary}. We also show that our claim remains valid when these approaches are used for synth-to-real semantic self-labeling. To show this, we rely on several synth-to-real unsupervised domain adaptation (UDA) methods. In particular, we use the HRDA \cite{hoyer2022hrda} and co-training \cite{gomez2023cotraining} frameworks. We show that adding {\thedataset} boosts semantic segmentation results to the state-of-the-art on synth-to-real UDA for semantic segmentation. On the other hand, we encourage researchers to not only use {\thedataset} for such a task but also for others requiring labels, for instance, segmentation, object detection, and depth estimation. Accordingly, we make {\thedataset} publicly available to the research community for free. In this regard, it is also important to note that, to generate images that can be used for training AI models, the digital assets used to create {\thedataset} have been carefully curated to ensure they are legally usable for that purpose.

The rest of the paper is organized as follows: Section \ref{sec:relatedwork} overviews the most related works, placing ours in context. Section \ref{sec:ourdataset} details the main components involved in the generation of {\thedataset}. Section \ref{sec:experimentalresults} shows the usefulness of {\thedataset} for synth-to-real UDA in semantic segmentation. Finally, Section \ref{sec:conclusions} summarizes the work presented in this paper and outlines related future work.


\begin{figure*}[hbt!]
    \centering
    \includegraphics[width=1\textwidth]{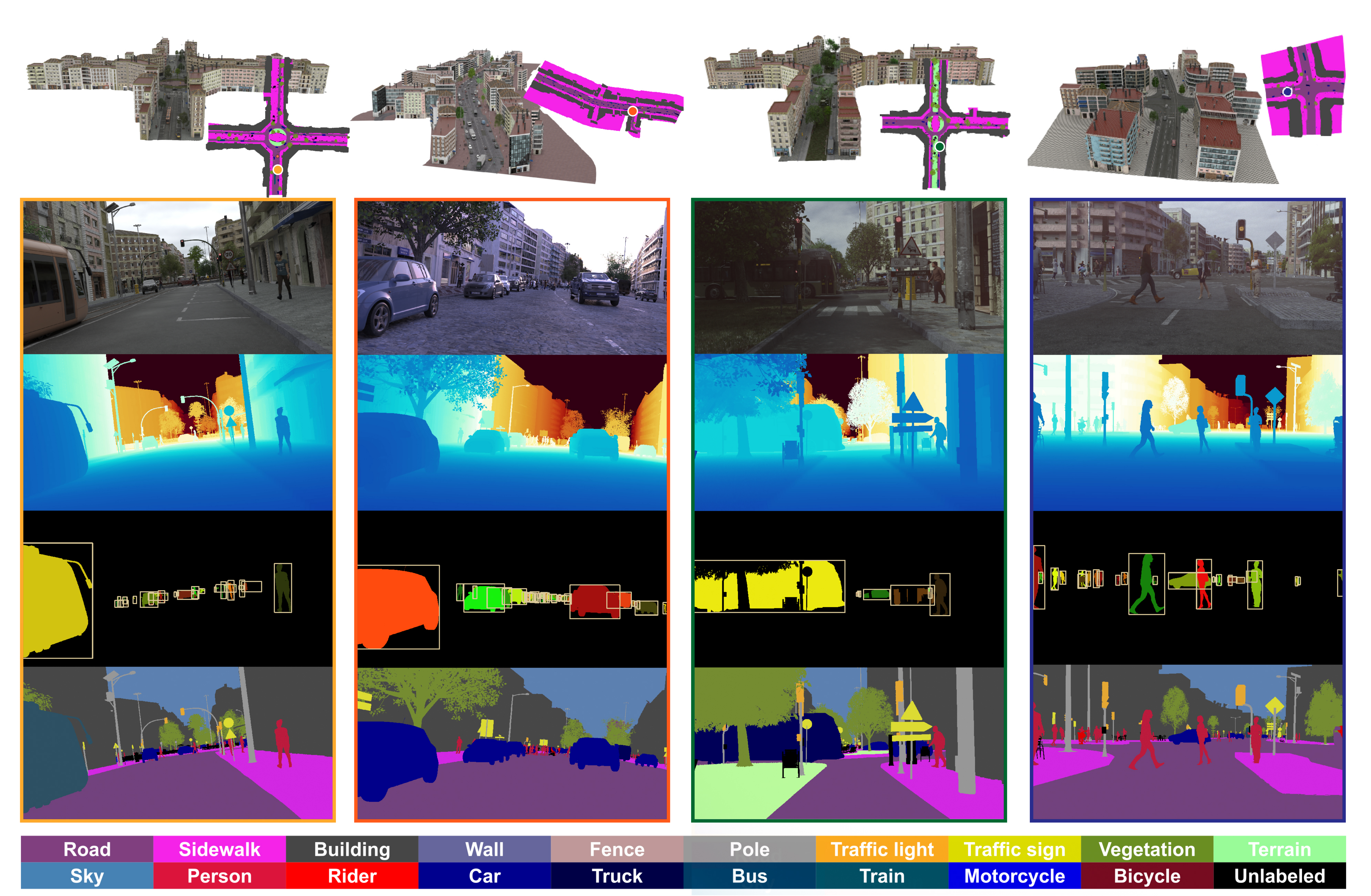}
    \caption{{\thedataset} covers $4$ different ODDs. It supports different lighting conditions, enabling atmospheric participating media ({\eg}, columns $3$ and $4$), procedurally generating different building landscapes, and shuffling locations and materials of assets. From top to bottom, we see a top view of the ODDs, RGB images captured around the indicated circles within the ODDs, corresponding depth maps (pseudo-color), object instances with their bounding boxes, and per-pixel class semantic labels.}
    \label{fig:dataset_overview}
\end{figure*}
\begin{figure*}[hbt!]
    \centering    
    \includegraphics[width=1\textwidth]{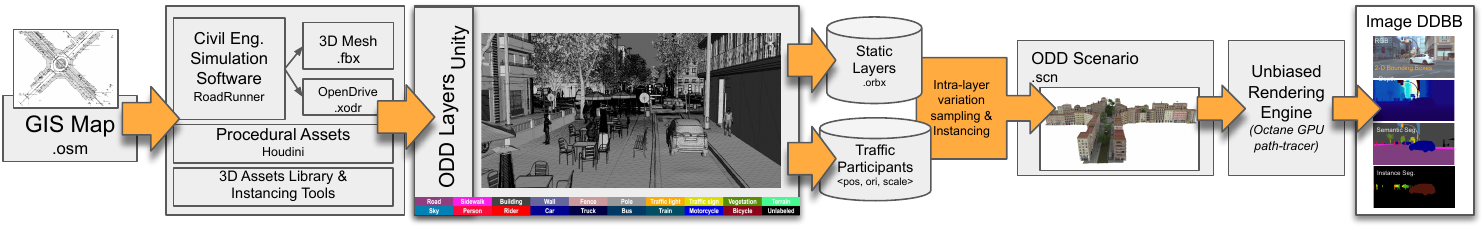}
    \caption{{\thedataset} content generation pipeline overview. Our builder scripts combine real GIS data with procedurally generated content and a 3D library of custom and commercial assets to create layers containing configurable content variations.}
    \label{fig:urbansyn_pipeline}
\end{figure*}

\section{Related work}
\label{sec:relatedwork}

\subsection{Synthetic datasets of driving scenarios.}

The use of synthetic datasets for visual machine learning is becoming increasingly popular for many tasks in computer vision, mostly motivated by its potential to reduce the need for expensive and time-consuming manual image labeling. A recent survey~\cite{tsirikoglou2020survey} provides a comprehensive overview of various image synthesis approaches according to their modeling and rendering procedures as well as their applications in computer vision. While a whole range of synthetic datasets exists for different domains providing various types of labeled information~\cite{shah2018airsim, roberts2021hypersim, fu20213dfuture, savva2019habitat}, we focus here on those presenting driving scenarios.

Using driving simulators as platforms to produce large amounts of automatically labeled images has been successfully proposed in the last decade~\cite{haltakov2013framework, ros2016synthia, richter2016playing, johnson2016driving, gaidon2016vkitti, dosovitskiy2017Carla, shah2018airsim}. However, while these approaches can rapidly generate thousands of labeled images, their results often lack a photorealistic level. This can be attributed to two factors: i) the overall quality used to model the underlying 3D assets, which often trades level of detail in exchange for simulation speed; and ii) the use of real-time game engines for rendering, which often employ trade-off implementations to approximate complex global illumination, reflections, shadows, and camera lens effects with less accuracy. Popular datasets generated under these trade-offs are GTAV~\cite{richter2016playing}, SYNTHIA~\cite{ros2016synthia}, Virtual KITTI~\cite{gaidon2016vkitti}, and those based on CARLA simulator \cite{dosovitskiy2017Carla}.

A recent trend to increase the realism of the generated images is to employ high-quality 3D assets together with offline Monte Carlo ray-tracing algorithms like path tracing~\cite{blasinski2018optimizing, wrenninge2018synscapes}. This was the case of Synscapes~\cite{wrenninge2018synscapes}, in which the gained photorealism came with an extra rendering cost. Another relevant aspect is the authoring cost employed in pursuing realistic layouts. Several works opted to incorporate procedural modeling in their pipelines; {\eg}, Synscapes incorporated a procedural scene generator, as well as ProcSy~\cite{khan2019procsy}, which delegated that task to CityEngine~\cite{parish2001procedural}. Another alternative consists of automatically synthesizing scene layouts tailored to solve downstream tasks, {\eg}, semantic segmentation. Meta-Sim~\cite{kar2019meta} aimed at automatically tuning parameters given a target collection of real images in an unsupervised way, while Meta-Sim 2~\cite{devaranjan2020metasim2} aims at learning the scene structure in addition to parameters. In both cases, automatically labeled images were generated through a graphics engine sharing trade-offs similar to those presented for most driving simulators.

\subsection{Synth-to-real semantic segmentation.}

Using synthetic images to train semantic segmentation models implies addressing the synth-to-real domain gap, which has attracted a significant amount of research in recent years. We advise readers to consult \cite{csurka2022semsegreview} for a comprehensive review. In this context, it is assumed that no human-provided labels are available for real-world images; thus, synth-to-real semantic segmentation becomes a problem of unsupervised domain adaptation (UDA).  

In this context, UDA methods are based on two general ideas, taken in isolation or combination. One idea is to perform a synthetic-real domain alignment of the input/features/output of the deep model under training. The other idea is to perform self-labeling ({\ie}, producing pseudo-labels) of the real-world images for their use in re-training/back-propagation loops. Attending to how coupled to the underlying deep model architecture is the implementation of these ideas, we find model-centric \cite{luo2019taking, qin2019generatively, choi2019self, tsai2019domain, tuan2019ADVENT, zhao2019multisource, Lv2020cross, kim2020learning, wang2020differential, wang2020classes, gao2021DSP, zheng2021rectifying, hoyer2022daformer, hoyer2022hrda, hong2024robust} and data-centric \cite{li2019bidirectional, zou2019confidence, subhani2020learning, chao2021rethinking, zhang2021multiple, he2021multisource, gomez2023cotraining} methods. Model-centric methods have been developed on top of a particular deep architecture (by accessing specific internal layers or modifying its loss function). Data-centric methods treat deep models pretty much as black boxes and mainly focus on assessing and guaranteeing the reliability of the predicted pseudo-labels. Model- and data-centric methods can be complemented with tailored data augmentation techniques \cite{tranheden2021DACS, olsson2021ClassMix, araslanov2021self}, with image-to-image translation \cite{zou2018unsupervised, chang2019allabout, yang2020FDA}, and with LAB color space alignment which, in line with findings regarding color spaces and robustness of deep models \cite{kanjar2021impact}, can meaningfully reduce the synth-to-real domain gap \cite{he2021multisource, gomez2023cotraining}. 

Most works use Cityscapes as real-world (target) domain, while SYNTHIA, GTAV, and/or Synscapes are used as synthetic (source) domains. The predominant experimental setting corresponds to using a single synthetic dataset as the source domain. However, some works have shown the convenience of relying on multiple synthetic sources \cite{gong2021mDALU, zhao2019multisource, he2021multisource, gomez2023cotraining} to reduce the synth-to-real domain gap, which is in agreement with findings in the pure real-world setting \cite{lambert2020Mseg}.

\subsection{In this paper.} 

We introduce {\thedataset}, a novel dataset for autonomous driving research generated by combining high-quality 3D assets, semi-procedural tools for scene layout generation, and path tracing rendering techniques to produce labeled images with a high degree of photorealism. By design, {\thedataset} is compact and brings differences from GTAV and Synscapes regarding road layouts, background content, and traffic participants. To show its usefulness, besides providing synthetic-only semantic segmentation baseline results, we use {\thedataset} for running synth-to-real UDA methods. We use the state-of-the-art HRDA \cite{hoyer2022hrda} to represent model-centric methods and the state-of-the-art co-training \cite{gomez2023cotraining} to represent data-centric methods. As the real-world target domain, our experiments not only consider Cityscapes as in most previous literature but also the more challenging BDD100K and Mapillary Vistas datasets. The obtained results confirm that both in isolation and as part of the Three Musketeers dataset, {\thedataset} fosters synth-to-real semantic segmentation. 


\section{{\thedataset}, the new musketeer} 
\label{sec:ourdataset}

This section summarizes different aspects concerning the generation of {\thedataset}, including the construction of our Operational Design Domains (ODDs) and some image rendering details. Overall, as novel real-world datasets are different from previous ones because they were acquired in different real-world areas using different camera sensors and lighting conditions, 
such differences also apply to {\thedataset} compared to GTAV and Synscapes, making them inherently complementary.

\subsection{Dataset features}
The {\thedataset} dataset provides photorealistic color images, paired with their ground-truth depth maps, pixel-level semantics and instance segmentation, as well as 2D object bounding boxes indicating their occlusion degree as a percentage value w.r.t. the whole object within the camera frustum. 
We adhere to the common $19$ classes of Cityscapes as the established de facto standard. Please check Figure \ref{fig:dataset_overview} to see some image examples and their corresponding labeled information.

{\thedataset} aims to model real-world imperfections affecting both the geometry and the material appearance of its assets.
In contrast to GTAV, we utilize unbiased path tracing instead of videogame deferred rendering. When compared to Synscapes, we enrich realism by simulating atmospheric effects like urban air pollution through physics-based light-transport simulation in participating media. In addition, we exploit procedural assets with randomized geometries and materials to enhance the variability of certain assets ({\eg}, buildings), and we also simulate different types of vehicle lights, all intending to 
elevate {\thedataset} to a new level of photorealism.

Moreover, {\thedataset} has been meticulously generated to optimize diversity and training value, balancing efficiency and quality and resulting in a more compact dataset than its predecessors. In particular, we showcase distinctive scene configurations, assets, and illumination profiles while minimizing redundant images as much as possible ({\eg}, we avoid capturing similar views under similar environment conditions). As a consequence, {\thedataset} consists of $7,539$ images, making it easy to share and store and being aligned with the \emph{train on less data} tip for decarbonizing AI development \cite{Caballar2024TheWay}. As we will see in Section \ref{sec:experimentalresults}, despite its compactness, {\thedataset} gives rise to the most accurate models for image semantic segmentation.

\subsection{Data generation pipeline}

\begin{figure}[t]
    \centering
    \includegraphics[width=\columnwidth]{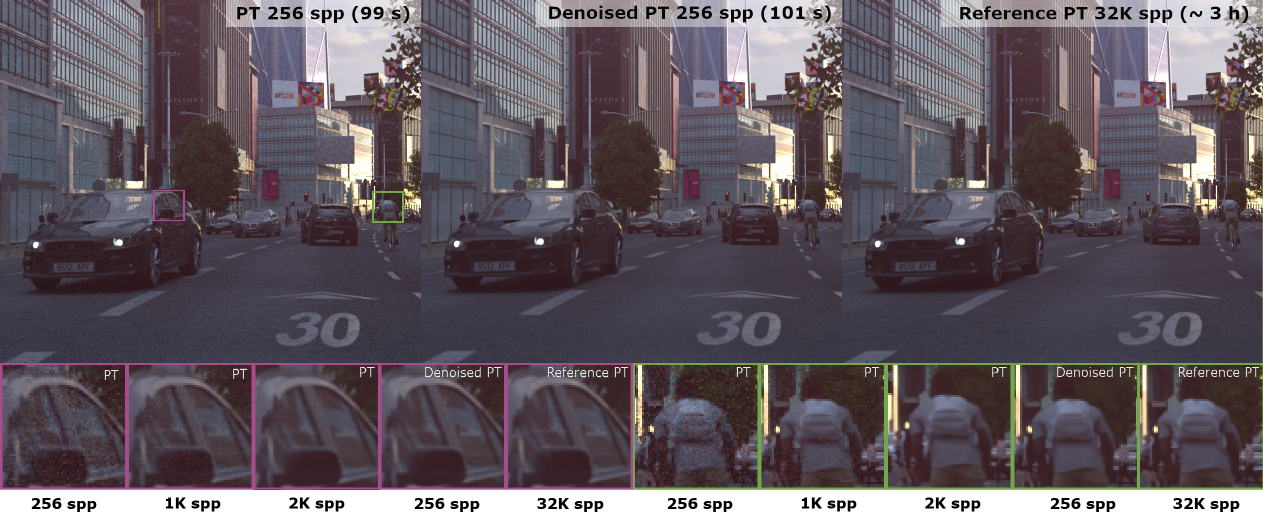}
    \caption{Rendering results using unbiased path tracing (PT) with different sample counts. We set the adaptive sampling to use a maximum of $256$ spp. Adaptive sampling and denoising help reduce undesired PT noise and large rendering times.
    }
    \label{fig:rendering_comparison}
\end{figure}

The designed pipeline to generate {\thedataset}, summarized in Figure~\ref{fig:urbansyn_pipeline}, allowed us to seamlessly integrate highly realistic and curated scene layouts with procedurally-generated content based on high-quality assets and materials.
Similarly to Synscapes or ProcSy, our dataset is mostly procedurally generated. However, our approach uses real maps from GIS sources, {\ie}, OpenStreetMaps (OSM), to model realistic scenario layouts. The main roads, lanes, and sidewalks are traced over using Mathworks RoadRunner, as well as potential locations of traffic signs or traffic lights. Then, a parameterized in-house builder script parses and instantiates a 3D variation of a given scenario layout inside the Unity game engine. We mainly utilize Unity as our simulation engine to verify plausible spatial occupancy among all assets, {\eg}, to avoid geometry collisions.
The system allows controlling per-class picking distributions in a statistically meaningful manner and supports per-asset weighted probabilities.
Our approach assembles these scenarios by parts (layers), sampling from a predefined set of available 3D assets\footnote{These assets were developed in-house or acquired with licenses that allow {\thedataset} redistribution under the CC-BY-SA 4.0 license.}, and automatically instancing procedural variations developed via Unity scripting and SideFX Houdini.

\subsection{Rendering details}
\label{ssec:rendering_details}
The images presented in {\thedataset} were generated using OctaneRender's implementation of unbiased path tracing~\cite{kajiya1986rendering}, a widely used technique in computer graphics and the film industry~\cite{keller2015path}. However, one of the well-known side effects of Monte Carlo (MC) path tracing is its variance noise~\cite{zwicker2015recent}. 

To generate {\thedataset}, we decided to employ adaptive sampling, a technique that adaptively dedicates more samples in those regions where the estimated MC variance noise is higher. We established a threshold of $128$ samples-per-pixel (spp) to estimate the initial per-pixel variance and let the adaptive sampling algorithm decide where to continue calculating more samples for each pixel.
To further reduce noise, we relied on OctaneRender's built-in deep learning-based denoising~\cite{huo2021survey}.
This allowed us to significantly reduce the total rendering time per frame, from potentially several hours to just a hundred seconds, all at the expense of trading MC variance noise with some minimal bias introduced by the denoising. The overhead of the denoiser is almost negligible, as it requires only from $1$ to $5$ seconds per image. Thus, the images of {\thedataset} were rendered and denoised using a maximum of $256$ spp. Figure~\ref{fig:rendering_comparison} illustrates how renders using $256$ spp and denoising could be perceptually very similar to unbiased PT renders with $\times 4$ to $\times 8$ more spp (\eg $1$K or $2$K spp). 

While the specific parameters of the path-tracing algorithm could be adjusted according to the target scenario, we utilized a set of predefined and recommended parameter values that worked well across the wide variety of scenarios covered by {\thedataset}. For example, the maximum amount of diffuse and specular (glossy) bounces was set to a maximum of $4$ bounces. The cost of rendering one frame also depends on the ODD.

\subsection{Operational Design Domains (ODDs)}
Using the aforementioned pipeline, we can generate different variations and asset distributions within our ODDs. For {\thedataset}, we have considered four, inspired by common real traffic layouts (see the top row of Figure~\ref{fig:dataset_overview}). Table~\ref{tab:odd_image_details} details the number of images, average, and total rendering time for each ODD. We can observe that using a participating medium has a huge impact on render times. Additionally, Table \ref{fig:ODDs} details the intrinsic variability of the ODDs, detailing the number of assets, materials, and variations for each class. We build the scenes used for rendering UrbanSyn, combining these assets to obtain the greatest amount of diversity while preserving the plausibility of the result.
To maintain sufficient control over random shuffling, we precomputed a series of intra-layer variations for static elements and conducted random sampling over them.
After that, dynamic traffic participants are positioned and later instantiated directly within the rendering engine. This alternative produced enough variations for our purpose of generating single images. 
%

The unbiased path tracer then takes the generated scenarios as input and produces the labeled images. Multiple steps are involved before generating the final RGB images, including tone-mapping and gamma transformations, to convert them to low-dynamic range (LDR) in the desired RGB color space.

\begin{table}[t]
\caption{UrbanSyn detailed characteristics per ODD. Rendered using two NVIDIA RTX 3090 GPUs in parallel.}
\label{tab:odd_image_details}
\centering
\scalebox{0.8}[0.8]{
\begin{tabular}{l|rcrr}
\#ODD &
  \begin{tabular}[c]{@{}r@{}}Num. of \\ images\end{tabular} &
  \begin{tabular}[c]{@{}c@{}}Use Part.\\ Media\end{tabular} &
  \begin{tabular}[c]{@{}r@{}}Average\\ rendering\\ per image\end{tabular} &
  \begin{tabular}[c]{@{}r@{}}Total\\ time\end{tabular} \\ \hline
ODD 1  & 1,319 & No            & 36.96s  & 13h 32m   \\
ODD 2  & 1,369 & Yes           & 120.82s  & 45h 56m    \\
ODD 3  & 2,905 & No            & 24.55s  & 19h 48m    \\
ODD 4  & 1,946 & Yes           & 87.95s  & 47h 32m     \\ \hline
Total & 7,539 & Y:43\%, N:57\% & 69,17s & 132h  46m \\ \hline
\end{tabular}
}
\end{table}

\begin{table*}
\caption{Details of the intrinsic variability we have implemented to capture the images composing {\thedataset}. Camera poses correspond to those commonly used onboard a regular car, either looking forward or backwards, with orientation variations.}
\label{fig:ODDs}
\centering
\centerline{\includegraphics[width=\textwidth]{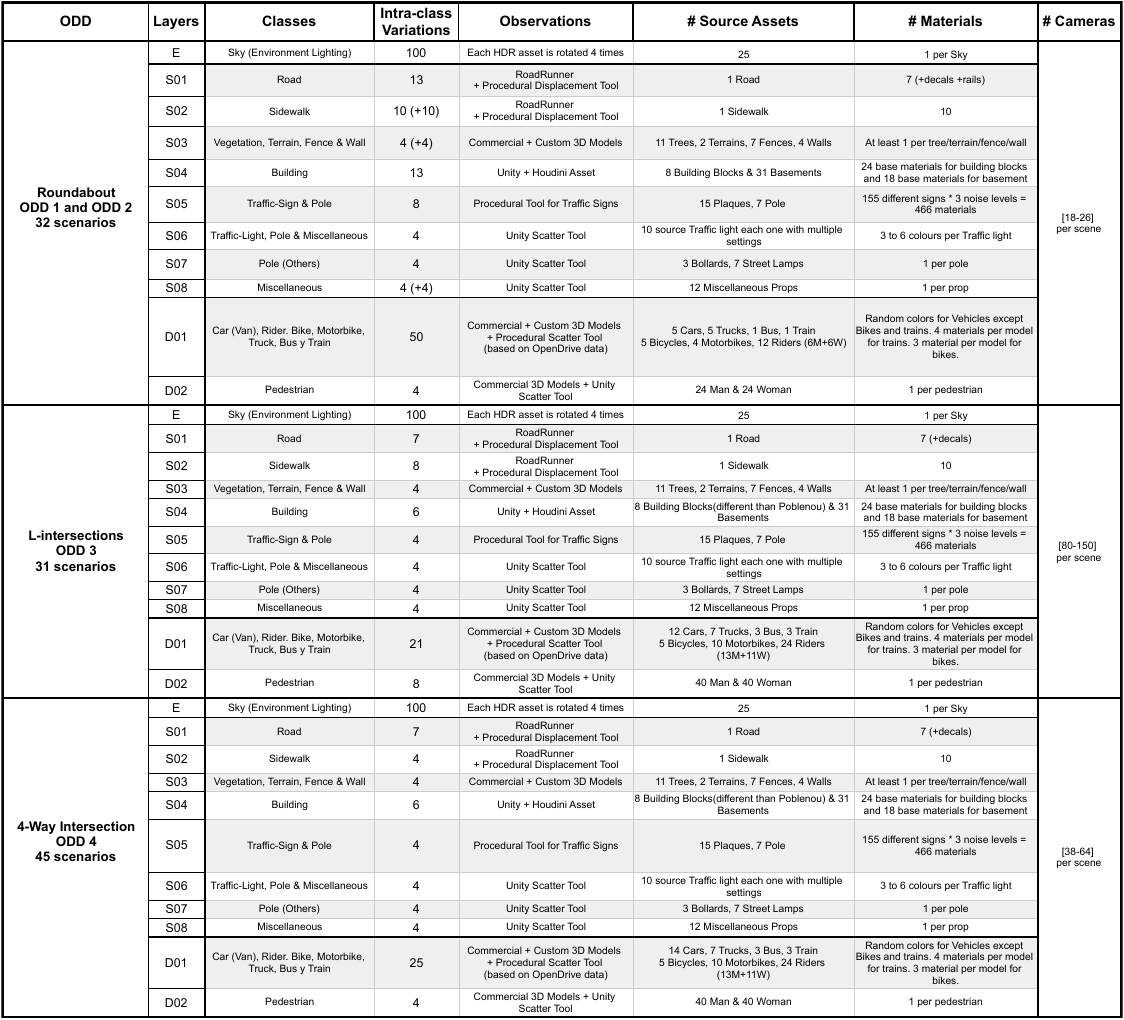}}
\end{table*}

\subsection{{\thedataset} brings diversity}
GTAV was acquired from a video game, bringing all the well-known limitations related to real-time simulation ({\eg}, limited degrees of graphics realism, and a limited number of object instances present at once in the scene). Additionally, GTAV suffers a clear bias towards American-style street maps and cars. Synscapes ~\textit{was designed to closely resemble the Cityscapes dataset in structure and content} as stated by their authors~\cite{wrenninge2018synscapes}. Unlike GTAV, the scenarios generated in Synscapes are biased toward German cities. It is based on small maps, making use of a limited variety of materials, simplified procedural roads and buildings, missing real-world imperfections, and thus striking CGI uncanny perfection. Moreover, images tend to be overpopulated for all classes lacking in realism on scene composition. On the other hand, Synscapes offered higher image quality by utilizing unbiased path-tracing rendering. 

\begin{figure*}[t]
    \centering
    \includegraphics[width=1\textwidth]{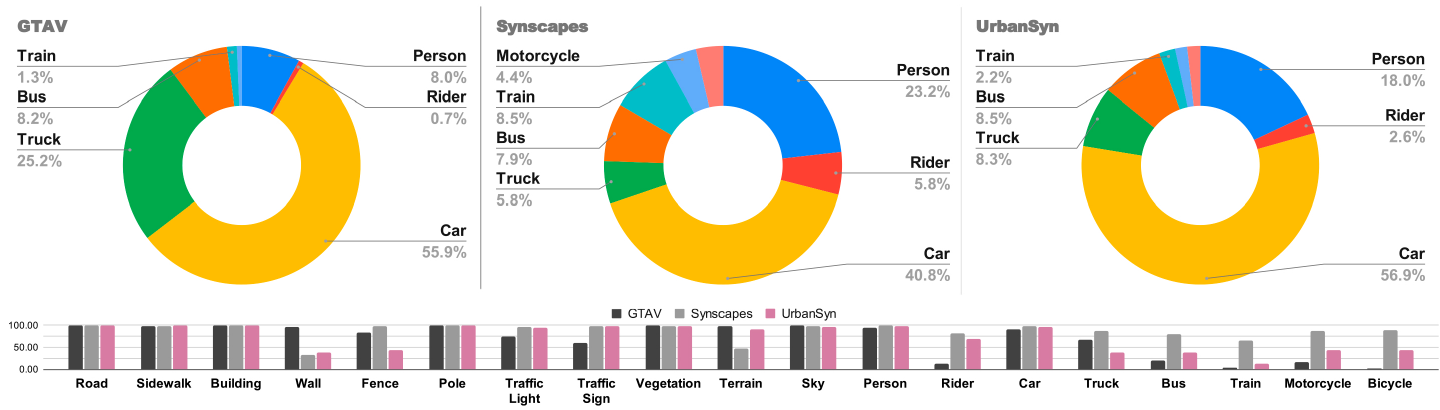}
    \caption{Content statistics for the Musketeers dataset: (Top) Per-class pixel-occupancy distributions. Apparently, the three datasets show relatively similar distributions, sharing more than what could set them apart. (Bottom) Percentage of images containing samples of the given class. {\thedataset} is on pair with Synscapes for certain classes, where both provide more examples than GTAV ({\eg}, Bus and Rider).}
    \label{fig:the3musketeers_pixeldata}
\end{figure*}

{\thedataset} has been conceived for diversity since its conception, not being tailored for any specific real-world dataset. Having in mind that more synthetic data does not necessarily guarantee a better model performance~\cite{Prakash2019structured}, we defined a set of clear requirements for the newly proposed dataset: i) it should achieve high graphics realism, at the level of Synscapes or beyond, to help to reduce synth-to-real visual domain shift; ii) the images should depict scenarios unseen in previous datasets, including real-world imperfections; iii) we aimed for semi-procedurally generated content seamlessly integrated with high-quality assets, so to get the better of both worlds in terms of realism and the covered variety. The achievement of these requirements has made {\thedataset} a great complement for GTAV and Synscapes. Thus, it makes sense to use {\thedataset} alone or as part of the Three Musketeers. Figure~\ref{fig:the3musketeers_pixeldata} details per-class statistics of pixel occupancy and percentage of images of the Three Musketeers. {\thedataset} is well-balanced and a trade-off between GTAV and Synscapes. On the one hand, it provides a more balanced variability and quantity than GTAV in classes such as Rider, Bus, Train, Motorcycle, and Bicycle. On the other hand, {\thedataset} maintains the scene realism and coherence, avoiding Synscapes practices to overpopulate all images always with every class on them ({\eg},~Synscapes have around 70\% of images that contain class Train, unrealistic in a real-world urban scenario.)

\newcommand{\lab}{\ell}
\newcommand{\ulab}{u}
\newcommand{\datasets}{\mathcal{X}}
\newcommand{\XL}{\datasets^{\lab}}
\newcommand{\XU}{\datasets^{\ulab}}
\newcommand{\XP}{\datasets^{\hat{\lab}}}
\newcommand{\synthetic}{\emph{\mbox{S}}}
\newcommand{\real}{\emph{\mbox{R}}}
\newcommand{\train}{\emph{\mbox{tr}}}
\newcommand{\test}{\emph{\mbox{tt}}}
\newcommand{\XLS}{\XL_\synthetic}
\newcommand{\XLRtr}{\XL_{{\real}_{\train}}}
\newcommand{\XLRtt}{\XL_{{\real}_{\test}}}
\newcommand{\XURtr}{\XU_{{\real}_{\train}}}
\newcommand{\XPRtr}{\XP_{{\real}_{\train}}}

\begin{figure*}[t]
    \centering    
    \includegraphics[width=1\textwidth]{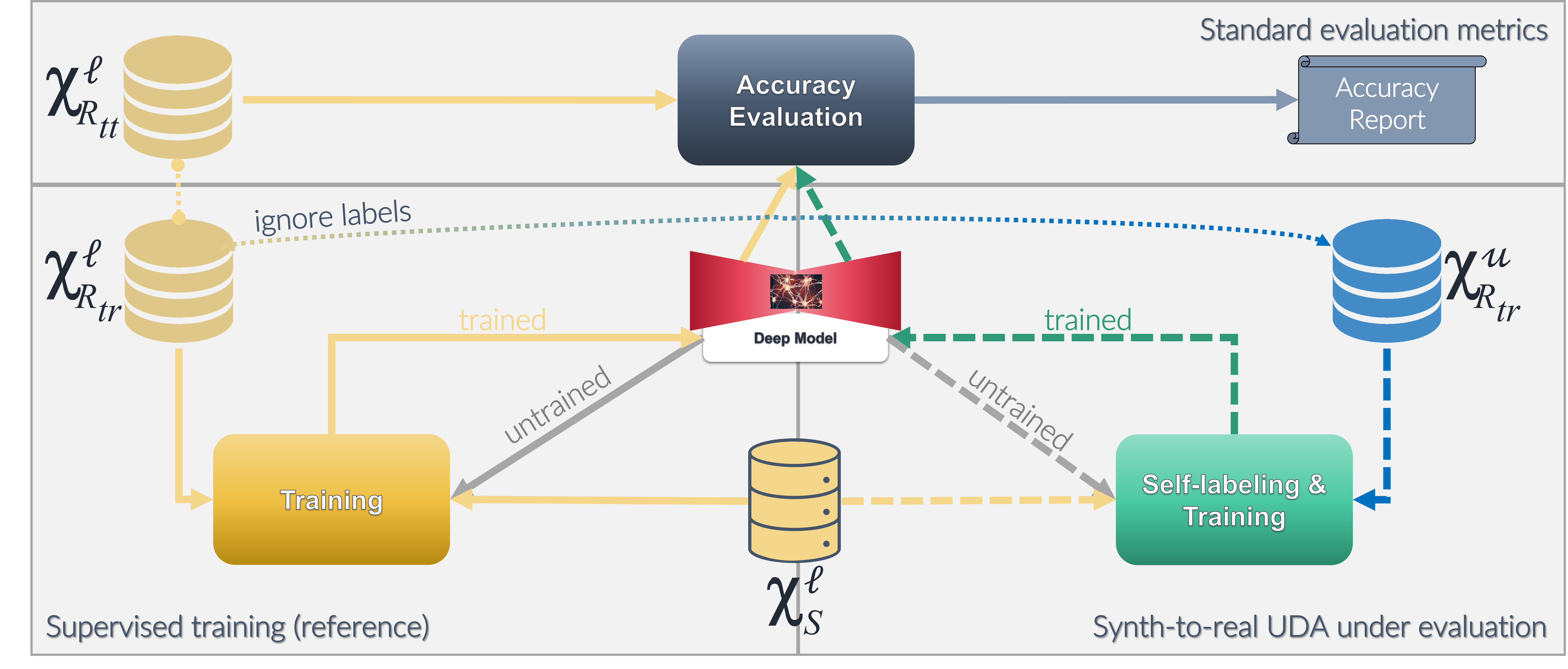}
    \caption{Experimental evaluation methodology for synth-to-real UDA procedures. Please, refer to the main text in section \ref{ssec:expemetho} for details.}
    \label{fig:evaluation_methodology}
\end{figure*}

\section{Experimental results}
\label{sec:experimentalresults}

In this paper, we highlight the usefulness of {\thedataset} by focusing on one of its ground truths, namely the pixel-level class labels. These labels are crucial for conducting synth-to-real UDA to develop semantic segmentation models without human labeling. We conduct comprehensive experiments in this section to demonstrate the importance of these labels.In the rest of this section, we begin by explaining the standard experimental methodology applied to evaluate synth-to-real UDA settings, for either assessing new UDA protocols or new synthetic datasets as is our purpose in this paper. Next, we detail the datasets and evaluation metrics used in our experiments. After, we provide additional implementation details. Finally, we present the experiments and discuss the results obtained. 

\subsection{Experimental methodology}
\label{ssec:expemetho}

Figure \ref{fig:evaluation_methodology} illustrates the standard experimental evaluation methodology used to assess synth-to-real UDA procedures, which we also adhere to in this paper. It is assumed that we have a deep model to be adapted, a source domain in the form of automatically labeled synthetic data, $\XLS$, and a target domain in the form of real-world unlabeled training data, $\XURtr$. In practice, to perform experiments, the real-world domain is composed of a training and a testing set, $\XLRtr$ and $\XLRtt$ respectively. These are used by the research community for different types of tasks. For instance, in the case of synth-to-real UDA, the set $\XLRtr$ plays the role of $\XURtr$ by ignoring its (human-provided) labels. Then, a standard training of the deep model using $\XLRtr$ is used as upper-bound reference for the synth-to-real UDA under evaluation; while an analogous training based on $\XLS$ is used as lower-bound reference. The goal of a synth-to-real UDA method is to use $\XLS$ and $\XURtr$ to obtain a similar (or better) performing deep model than if $\XLRtr$ is used as the training set. This process involves self-labeling $\XURtr$. To assess model accuracy irrespective of its training, $\XLRtt$ and a standard evaluation procedure are used. The assumption is that good synth-to-real UDA procedures will be able to address the training of deep models for new unlabeled real-world datasets at least using the same synthetic data or improved versions of it. 

This general methodology translates into specific experiments to assess the goodness of {\thedataset}. Table \ref{tab:essentialquestions} summarizes them. We run this table of experiments twice, {\ie}, to cover two significantly different deep models for semantic segmentation and their respective UDA procedures, namely, DeepLabV3+ (pure CNN-based model) and SegFormer MiT-B5 (Transformer-based model) with Co-training and HRDA respective UDA procedures. Comparing experiments $r-r-\mbox{Sup}$ with corresponding $s-r-\mbox{Sup}$, where $r\in\{\mbox{C,B,M}\}$ and $s\in\{\mbox{G,S,U,GS,GSU}\}$, we can see domain gaps in accuracy for the supervised settings. Note that in the $r-r-\mbox{Sup}$ experiments, real-world human-labeled images are used for training and there is no domain gap, thus, the resulting accuracy can be taken as an upper-bound reference. In the $s-r-\mbox{Sup}$ experiments, synthetic-only automatically labeled images are used for training and there is a domain gap with the real-world images used for testing, thus, the resulting accuracy can be taken as a lower-bound reference or baseline. We can see how large the domain gap between given $s$ and $r$ is by comparing the accuracy resulting from the $r-r-\mbox{Sup}$ and $s-r-\mbox{Sup}$ experiments. In the UDA setting, we can see the contribution of the synthetic data $s$ by comparing the accuracy of experiments $s-r-\mbox{UDA}$ to $s-r-\mbox{Sup}$ and $r-r-\mbox{Sup}$, varying $r$ and the UDA method, where the aim is to approach (or surpass) the accuracy of $r-r-\mbox{Sup}$ after significantly improving $s-r-\mbox{Sup}$. If we restrict $s$ to a single synthetic dataset (here $\{\mbox{G,S,U}\}$), we can compare their individual contributions. If $s$ comes from the combination of different synthetic datasets, we can evaluate if they are complementary to the UDA task. 

According to this standard setting of experiments, we will see in section \ref{ssec:resana} that {\thedataset} allows to significantly increase the accuracy of different state-of-the-art synth-to-real UDA procedures for semantic segmentation (DeepLabV3+/Co-training, SegFormer MiT-B5/HRDA), in single-source regime--individually comparing {\thedataset} (U) to GTAV (G) and Synscapes (S)--and in a multi-source regime--comparing the use of GTAV+Synscapes+{\thedataset} (GSU) to only using GTAV+Synscapes (GS)--, and working with different real-world target domains--Cityscapes (C), BDD10K (B), Mapillary Vistas (M)--.

\begin{table}
\caption{Types of experiments to be carried out. Each experiment ID gives rise to experiments based on different deep architectures for semantic segmentation, {\ie}, DeepLabV3+ and SegFormer in the supervised settings (Sup), including the respective Co-training and HRDA procedures in the UDA settings. For simplicity, we do not explicitly indicate these models and procedures in this table. We also indicate in which table of the paper the reader can find the respective experiments. Please, refer to the main text in section \ref{ssec:expemetho} for details.}
\label{tab:essentialquestions}
\centering
{\scriptsize
\scalebox{0.9}[0.9]{
\begin{tabular}{r|c|c|c|c|c}
\noalign{\hrule height 2pt}
Experiment & \multirow{2}{*}{$\XLS$}    & \multirow{2}{*}{$\XLRtr$ from ...} & \multirow{2}{*}{$\XURtr$ from ...} & \multirow{2}{*}{$\XLRtt$ from ...} & \multirow{2}{*}{Where?}                 \\ 
Type ID   &                             &                                    &                                    &                                    &                                         \\ 
\noalign{\hrule height 2pt}
\rowcolor{gray!20} \multicolumn{6}{c}{Real-to-real supervised training (same domain)}        \\
\hline
C-C-Sup    & \multirow{3}{*}{-}         &  Cityscapes                        & \multirow{3}{*}{-}                 & Cityscapes                         & Table \ref{tab:cityscapes_quantitative} \\ 
\cline{1-1} \cline{3-3} \cline{5-6}
B-B-Sup    &                            &  BDD100K                           &                                    & BDD100K                            & Table \ref{tab:bdd_quantitative}        \\
\cline{1-1} \cline{3-3} \cline{5-6}
M-M-Sup    &                            &  Mapillary                         &                                    & Mapillary                          & Table \ref{tab:mapillary_quantitative}  \\ 
\hline
\noalign{\hrule height 1pt}
\rowcolor{gray!20} \multicolumn{6}{c}{Synth-to-real supervised training (domain gap, no UDA)} \\
\hline
G-C-Sup    & \multirow{3}{*}{GTAV}      & \multirow{3}{*}{-}                 & \multirow{3}{*}{-}                 & Cityscapes                         & Table \ref{tab:cityscapes_quantitative} \\ 
\cline{1-1} \cline{5-6}
G-B-Sup    &                            &                                    &                                    & BDD100K                            & Table \ref{tab:bdd_quantitative}        \\ 
\cline{1-1} \cline{5-6}
G-M-Sup    &                            &                                    &                                    & Mapillary                          & Table \ref{tab:mapillary_quantitative}  \\ 
\hline
S-C-Sup    & \multirow{3}{*}{Synscapes} & \multirow{3}{*}{-}                 & \multirow{3}{*}{-}                 & Cityscapes                         & Table \ref{tab:cityscapes_quantitative} \\ 
\cline{1-1} \cline{5-6}
S-B-Sup    &                            &                                    &                                    & BDD100K                            & Table \ref{tab:bdd_quantitative}        \\ 
\cline{1-1} \cline{5-6}
S-M-Sup    &                            &                                    &                                    & Mapillary                          & Table \ref{tab:mapillary_quantitative}  \\ 
\hline
U-C-Sup    & \multirow{3}{*}{UrbanSyn}  & \multirow{3}{*}{-}                 & \multirow{3}{*}{-}                 & Cityscapes                         & Table \ref{tab:cityscapes_quantitative} \\ 
\cline{1-1} \cline{5-6}
U-B-Sup    &                            &                                    &                                    & BDD100K                            & Table \ref{tab:bdd_quantitative}        \\ 
\cline{1-1} \cline{5-6}
U-M-Sup    &                            &                                    &                                    & Mapillary                          & Table \ref{tab:mapillary_quantitative}  \\ 
\hline
GS-C-Sup   & GTAV                       & \multirow{3}{*}{-}                 & \multirow{3}{*}{-}                 & Cityscapes                         & Table \ref{tab:cityscapes_quantitative} \\ 
\cline{1-1} \cline{5-6}
GS-B-Sup   & +                          &                                    &                                    & BDD100K                            & Table \ref{tab:bdd_quantitative}        \\ 
\cline{1-1} \cline{5-6}
GS-M-Sup   & Synscapes                  &                                    &                                    & Mapillary                          & Table \ref{tab:mapillary_quantitative}  \\ 
\hline
GSU-C-Sup  & GTAV                       & \multirow{3}{*}{-}                 & \multirow{3}{*}{-}                 & Cityscapes                         & Table \ref{tab:cityscapes_quantitative} \\ 
\cline{1-1} \cline{5-6}
GSU-B-Sup  & + Synscapes                &                                    &                                    & BDD100K                            & Table \ref{tab:bdd_quantitative}        \\ 
\cline{1-1} \cline{5-6}
GSU-M-Sup  & + Urbansyn                 &                                    &                                    & Mapillary                          & Table \ref{tab:mapillary_quantitative}  \\ 
\noalign{\hrule height 1pt}
\rowcolor{gray!20} \multicolumn{6}{c}{Synth-to-real UDA training (domain gap, self-labeling based UDA)} \\
\hline
G-C-UDA   & \multirow{3}{*}{GTAV}       & \multirow{3}{*}{-}                 & Cityscapes                         & Cityscapes                         & Table \ref{tab:cityscapes_quantitative} \\ 
\cline{1-1} \cline{4-6}
G-B-UDA   &                             &                                    & BDD100K                            & BDD100K                            & Table \ref{tab:bdd_quantitative}        \\ 
\cline{1-1} \cline{4-6}
G-M-UDA   &                             &                                    & Mapillary                          & Mapillary                          & Table \ref{tab:mapillary_quantitative}  \\ 
\hline
S-C-UDA   & \multirow{3}{*}{Synscapes}  & \multirow{3}{*}{-}                 & Cityscapes                         & Cityscapes                         & Table \ref{tab:cityscapes_quantitative} \\ 
\cline{1-1} \cline{4-6}
S-B-UDA   &                             &                                    & BDD100K                            & BDD100K                            & Table \ref{tab:bdd_quantitative}        \\ 
\cline{1-1} \cline{4-6}
S-M-UDA   &                             &                                    & Mapillary                          & Mapillary                          & Table \ref{tab:mapillary_quantitative}  \\ 
\hline
U-C-UDA   & \multirow{3}{*}{UrbanSyn}   & \multirow{3}{*}{-}                 & Cityscapes                         & Cityscapes                         & Table \ref{tab:cityscapes_quantitative} \\ 
\cline{1-1} \cline{4-6}
U-B-UDA   &                             &                                    & BDD100K                            & BDD100K                            & Table \ref{tab:bdd_quantitative}        \\ 
\cline{1-1} \cline{4-6}
U-M-UDA   &                             &                                    & Mapillary                          & Mapillary                          & Table \ref{tab:mapillary_quantitative}  \\ 
\hline
GSU-C-UDA & GTAV                        & \multirow{3}{*}{-}                 & Cityscapes                         & Cityscapes                         & Table \ref{tab:cityscapes_quantitative} \\ 
\cline{1-1} \cline{4-6}
GSU-B-UDA & + Synscapes                 &                                    & BDD100K                            & BDD100K                            & Table \ref{tab:bdd_quantitative}        \\ 
\cline{1-1} \cline{4-6}
GSU-M-UDA & + Urbansyn                  &                                    & Mapillary                          & Mapillary                          & Table \ref{tab:mapillary_quantitative}  \\ 
\noalign{\hrule height 2pt}
\end{tabular}
}
}
\end{table}

\subsection{Datasets and evaluation}
\label{ssec:data-eval}

The source (synthetic) datasets in common use to address synth-to-real UDA for semantic segmentation are GTAV \cite{richter2016playing}, Synscapes \cite{wrenninge2018synscapes}, and SYNTHIA \cite{ros2016synthia}. GTAV and Synscapes have semantic labeling for the 19 classes of Cityscapes \cite{cordts2016cityscapes}, while this is not the case for SYNTHIA. Thus, since we are especially interested in a multi-source setting, in this paper, we define the Three Musketeers dataset as consisting of GTAV, Synscapes, and {\thedataset}. GTAV is composed of 24,966 images with a resolution of $1914\times1052$ pixels directly obtained from the render engine of the videogame GTA V. Synscapes is composed of 25,000 images with a resolution of $1440\times720$ pixels of urban scenes, obtained by using a physics-based rendering pipeline. {\thedataset} is composed of 7,539 images with a resolution of $2048\times1024$ pixels. 

As target (real-world) datasets, we not only consider Cityscapes \cite{cordts2016cityscapes} as in most previous literature but also the more challenging BDD100K \cite{yu2020BDD} and Mapillary Vistas \cite{neuhold2017mapillary} datasets. Cityscapes is well-known for its semantic segmentation challenge, comprising 2,975 images for training and 500 for validation. The images have a resolution of $2048\times1024$ pixels and were obtained from different cities in Germany. BDD100K contains images taken from different vehicles in different USA cities. It is divided into 7,000 images for training and 1,000 for validation, with a resolution of $1280\times720$ pixels. As in \cite{gomez2023cotraining}, for training purposes, we only use daytime images without heavy occlusions and favorable weather, resulting in a subset of 1,777 images. We consider the validation set as it comes. Mapillary Vistas is composed of a large number of images from around the world. These images were obtained with different camera devices, generating a large variety of resolutions and aspect ratios. Some images were captured onboard, and others were taken from personal devices at the street level. As in \cite{gomez2023cotraining}, we only consider images with an aspect ratio of 4:3, which comprises more than 75\% of all the dataset images. This results in 14,716 images for training and 1,617 for validation. 

For each model that we train, we assess its accuracy in the validation set of each real-world dataset according to the 19 official classes defined for Cityscapes. As is common practice, other classes are ignored. Model training in the synth-to-real UDA setting is performed by considering the synthetic datasets and the training set of each real-world dataset. The semantic labels of these real-world training images are ignored. These labels are just considered to train reference models, {\ie}, those trained and validated on the same real-world dataset domain. As is standard, to assess segmentation accuracy, we use the intersection-over-union metric \cite{everingham2015Pascal}, {\ie}, $IoU = TP/(TP + FP + FN)$, where TP, FP, and FN refer to true positives, false positives, and false negatives, respectively. Given a pixel $p$ with ground truth class $c_p$ and predicted class $\hat{c}_p$, if $c_p=\hat{c}_p$ it counts as TP for class $c$, if $c_p\neq\hat{c}_p$ it counts as FN for class $c_p$ and as FP for class $\hat{c}_p$. Since IoU is computed per class, their mean (mIoU) is used to consider all the classes at once. We repeat each experiment three times and report its average and standard deviation. 

\begin{table*}[t]
\vspace{-4mm}
\caption{Target: Cityscapes. Per-class IoU and global mIoU values are shown. For each row, we report the mean and standard deviation based on three train-test runs. For each column, we highlight the best mean IoU/mIoU per \textbf{sub-block} (bold) and \underline{\textbf{block}} (bold-underlined). Two means are highlighted as best if their difference is $<1$.}
\label{tab:cityscapes_quantitative}
\centering
\setlength{\tabcolsep}{0.9mm} 

\scalebox{0.8}[0.8]{
\begin{tabular}{l|c|c|ccccccccccccccccccc!{\vrule width 1pt}c!{\vrule width 1pt}}

Method & Source & & \rotatebox[origin=c]{90}{\footnotesize Road} & \rotatebox[origin=c]{90}{\footnotesize Sidewalk} & \rotatebox[origin=c]{90}{\footnotesize Building} & \rotatebox[origin=c]{90}{\footnotesize Wall} & \rotatebox[origin=c]{90}{\footnotesize Fence} & \rotatebox[origin=c]{90}{\footnotesize Pole} & \rotatebox[origin=c]{90}{\footnotesize Traffic Light} & \rotatebox[origin=c]{90}{\footnotesize Traffic Sign} & \rotatebox[origin=c]{90}{\footnotesize Vegetation} & \rotatebox[origin=c]{90}{\footnotesize Terrain} & \rotatebox[origin=c]{90}{\footnotesize Sky} & \rotatebox[origin=c]{90}{\footnotesize Person} & \rotatebox[origin=c]{90}{\footnotesize Rider} & \rotatebox[origin=c]{90}{\footnotesize Car} & \rotatebox[origin=c]{90}{\footnotesize Truck} & \rotatebox[origin=c]{90}{\footnotesize Bus} & \rotatebox[origin=c]{90}{\footnotesize Train} & \rotatebox[origin=c]{90}{\footnotesize Motorbike} & \rotatebox[origin=c]{90}{\footnotesize Bike} & mIoU  \\ 

\noalign{\hrule height 2pt}

\multirow{2}{*}{SegFormer} & \multirow{2}{*}{GTAV}  & mean & $76.9$  & $27.7$  & $85.0$  & $29.5$  & $34.5$  & $38.2$  & $54.5$  & $21.5$  & $86.8$  & $36.8$  & $84.2$  & $69.9$  & $30.2$  & $79.4$  & $35.5$  & $39.3$  & $18.5$  & $28.3$  & $30.7$  & $47.8$ \\ 
& & std & $\stdx{1.4}$  & $\stdx{3.3}$  & $\stdx{0.2}$  & $\stdx{1.9}$  & $\stdx{3.0}$  & $\stdx{1.0}$  & $\stdx{0.7}$  & $\stdx{3.1}$  & $\stdx{0.6}$  & $\stdx{1.5}$  & $\stdx{0.4}$  & $\stdx{0.6}$  & $\stdx{1.1}$  & $\stdx{7.9}$  & $\stdx{2.4}$  & $\stdx{2.3}$  & $\stdx{9.4}$  & $\stdx{5.2}$  & $\stdx{2.7}$  & $\stdx{1.2}$ \\
\multirow{2}{*}{SegFormer} & \multirow{2}{*}{Synscapes}   & mean & $\bbest{91.9}$  & $\bbest{49.4}$  & $78.9$  & $\bbest{36.4}$  & $29.4$  & $50.3$  & $58.1$  & $61.9$  & $\bbest{86.9}$  & $\bbest{38.1}$  & $84.5$  & $72.2$  & $39.7$  & $89.8$  & $25.1$  & $18.7$  & $16.4$  & $38.9$  & $64.8$  & $54.3$ \\ 
& & std & $\stdx{0.9}$  & $\stdx{2.5}$  & $\stdx{3.0}$  & $\stdx{2.9}$  & $\stdx{3.8}$  & $\stdx{1.0}$  & $\stdx{1.3}$  & $\stdx{1.0}$  & $\stdx{0.7}$  & $\stdx{2.8}$  & $\stdx{3.2}$  & $\stdx{0.8}$  & $\stdx{1.1}$  & $\stdx{0.8}$  & $\stdx{2.0}$  & $\stdx{5.5}$  & $\stdx{0.5}$  & $\stdx{2.8}$  & $\stdx{1.7}$  & $\stdx{1.2}$ \\ 
\multirow{2}{*}{SegFormer} & \multirow{2}{*}{{\thedataset}}   & mean & $89.8$  & $46.1$  & $\bbest{87.3}$  & $21.6$  & $\bbest{47.6}$  & $\bbest{56.4}$  & $\bbest{64.0}$  & $\bbest{69.6}$  & $\bbest{87.2}$  & $32.6$  & $\bbest{88.5}$  & $\bbest{77.2}$  & $\bbest{54.9}$  & $\bbest{93.1}$  & $\bbest{69.1}$  & $\bbest{65.3}$  & $\bbest{28.5}$  & $\bbest{50.0}$  & $\bbest{73.2}$  & $\bbest{63.3}$ \\ 
& & std & $\stdx{0.3}$  & $\stdx{1.1}$  & $\stdx{0.2}$  & $\stdx{5.6}$  & $\stdx{0.7}$  & $\stdx{0.4}$  & $\stdx{0.4}$  & $\stdx{0.6}$  & $\stdx{0.4}$  & $\stdx{3.4}$  & $\stdx{1.0}$  & $\stdx{0.8}$  & $\stdx{0.9}$  & $\stdx{0.4}$  & $\stdx{4.9}$  & $\stdx{3.0}$  & $\stdx{10.2}$  & $\stdx{1.5}$  & $\stdx{0.3}$  & $\stdx{0.7}$ \\  \hline

\multirow{2}{*}{SegFormer} & \multirow{2}{*}{GTAV+Synscapes}  & mean & $90.5$  & $50.9$  & $\bbest{89.7}$  & $47.9$  & $47.4$  & $54.0$  & ${65.5}$  & $54.8$  & $\bbest{89.8}$  & ${47.4}$  & $\bbest{92.9}$  & $74.0$  & $37.4$  & $90.6$  & $46.2$  & $64.3$  & $\bbest{61.3}$  & $52.6$  & $62.9$  & $64.2$ \\ 
& & std & $\stdx{1.2}$  & $\stdx{3.0}$  & $\stdx{0.1}$  & $\stdx{2.3}$  & $\stdx{0.7}$  & $\stdx{0.4}$  & $\stdx{0.2}$  & $\stdx{1.9}$  & $\stdx{0.4}$  & $\stdx{2.9}$  & $\stdx{0.2}$  & $\stdx{0.4}$  & $\stdx{3.4}$  & $\stdx{0.5}$  & $\stdx{2.9}$  & $\stdx{1.6}$  & $\stdx{5.8}$  & $\stdx{3.2}$  & $\stdx{2.3}$  & $\stdx{0.8}$ \\

\multirow{2}{*}{SegFormer} & \multirow{2}{*}{GTAV+Synscapes+{\thedataset}}   & mean & $\bbest{92.3}$  & $\bbest{57.7}$  & $\bbest{90.3}$  & $\bbest{50.2}$  & $\bbest{53.2}$  & $\bbest{58.1}$  & $\bbest{66.7}$  & $\bbest{67.8}$  & $\bbest{90.0}$  & $\bbest{48.5}$  & $91.0$  & $\bbest{79.8}$  & $\bbest{56.5}$  & $\bbest{93.2}$  & $\bbest{71.3}$  & $\bbest{77.1}$  & $60.3$  & $\bbest{61.4}$  & $\bbest{73.5}$  & $\bbest{70.5}$ \\ 
& & std & $\stdx{0.1}$  & $\stdx{0.8}$  & $\stdx{0.3}$  & $\stdx{0.5}$  & $\stdx{0.9}$  & $\stdx{0.6}$  & $\stdx{0.2}$  & $\stdx{1.2}$  & $\stdx{0.1}$  & $\stdx{0.4}$  & $\stdx{1.2}$  & $\stdx{0.5}$  & $\stdx{1.7}$  & $\stdx{0.6}$  & $\stdx{1.4}$  & $\stdx{2.0}$  & $\stdx{4.7}$  & $\stdx{2.9}$  & $\stdx{0.4}$  & $\stdx{0.4}$ \\  \hline

\multirow{2}{*}{HRDA} & \multirow{2}{*}{GTAV}  & mean & $90.3$  & $32.1$  & $\gbbest{91.1}$  & $\gbbest{60.0}$  & $52.8$  & $58.2$  & $64.7$  & ${71.9}$  & $\gbbest{91.5}$  & $48.2$  & $\gbbest{94.1}$  & $75.0$  & $22.9$  & $94.0$  & ${83.2}$  & $73.4$  & $19.3$  & $65.2$  & $66.5$  & $66.0$ \\ 
\UDAbackbone{(DAFormer*)} & & std & $\stdx{3.8}$  & $\stdx{27.4}$  & $\stdx{0.1}$  & $\stdx{1.3}$  & $\stdx{2.2}$  & $\stdx{1.2}$  & $\stdx{0.2}$  & $\stdx{0.5}$  & $\stdx{0.1}$  & $\stdx{1.3}$  & $\stdx{0.1}$  & $\stdx{2.7}$  & $\stdx{21.0}$  & $\stdx{0.1}$  & $\stdx{0.9}$  & $\stdx{8.2}$  & $\stdx{27.3}$  & $\stdx{0.5}$  & $\stdx{0.7}$  & $\stdx{4.9}$ \\
\multirow{2}{*}{HRDA} & \multirow{2}{*}{Synscapes}  & mean & $\bbest{93.6}$  & $\bbest{70.6}$  & $89.1$  & $45.9$  & $49.5$  & $\gbbest{62.6}$  & $\gbbest{68.5}$  & $64.6$  & $90.5$  & $\bbest{51.7}$  & $\gbbest{95.0}$  & $77.8$  & $39.4$  & $93.5$  & $63.8$  & $47.3$  & $65.3$  & $53.9$  & $\gbbest{76.2}$  & $68.4$ \\ 
\UDAbackbone{(DAFormer*)} & & std & $\stdx{1.1}$  & $\stdx{3.7}$  & $\stdx{1.3}$  & $\stdx{1.7}$  & $\stdx{9.9}$  & $\stdx{0.4}$  & $\stdx{1.1}$  & $\stdx{3.5}$  & $\stdx{0.3}$  & $\stdx{3.3}$  & $\stdx{0.0}$  & $\stdx{0.5}$  & $\stdx{2.5}$  & $\stdx{0.6}$  & $\stdx{3.8}$  & $\stdx{7.2}$  & $\stdx{6.7}$  & $\stdx{10.4}$  & $\stdx{0.2}$  & $\stdx{0.9}$ \\
\multirow{2}{*}{HRDA} & \multirow{2}{*}{{\thedataset}}   & mean & $90.5$  & $56.1$  & $\gbbest{91.5}$  & $\gbbest{60.4}$  & $\gbbest{56.9}$  & $\bbest{61.7}$  & $66.2$  & $\bbest{73.5}$  & $\bbest{90.9}$  & $46.4$  & $\gbbest{94.8}$  & $\gbbest{81.3}$  & $\gbbest{61.2}$  & $\gbbest{95.0}$  & $\gbbest{84.7}$  & $\bbest{76.5}$  & $38.1$  & ${63.9}$  & $\gbbest{75.8}$  & $\bbest{71.9}$ \\ 
\UDAbackbone{(DAFormer*)} & & std & $\stdx{0.1}$  & $\stdx{0.3}$  & $\stdx{0.3}$  & $\stdx{1.1}$  & $\stdx{0.6}$  & $\stdx{0.9}$  & $\stdx{0.2}$  & $\stdx{0.2}$  & $\stdx{0.2}$  & $\stdx{2.2}$  & $\stdx{0.1}$  & $\stdx{0.4}$  & $\stdx{0.1}$  & $\stdx{0.0}$  & $\stdx{3.0}$  & $\stdx{4.8}$  & $\stdx{13.2}$  & $\stdx{1.4}$  & $\stdx{0.4}$  & $\stdx{0.7}$ \\  \hline

\multirow{2}{*}{HRDA} & \multirow{2}{*}{GTAV+Synscapes+{\thedataset}}   & mean & $\gbbest{97.0}$  & $\gbbest{77.8}$  & $\gbbest{92.0}$  & $59.0$  & $\gbbest{57.0}$  & $\gbbest{63.1}$  & $\gbbest{68.4}$  & $\gbbest{75.3}$  & $\gbbest{91.9}$  & $\gbbest{55.5}$  & $\gbbest{94.9}$  & $\gbbest{81.4}$  & $58.4$  & $\gbbest{94.9}$  & $83.4$  & $\gbbest{88.4}$  & $\gbbest{75.4}$  & $\gbbest{65.7}$  & $\gbbest{76.1}$  & $\gbbest{76.6}$ \\ 
\UDAbackbone{(DAFormer*)} & & std & $\stdx{0.3}$  & $\stdx{1.6}$  & $\stdx{0.1}$  & $\stdx{0.7}$  & $\stdx{0.3}$  & $\stdx{0.3}$  & $\stdx{0.2}$  & $\stdx{0.3}$  & $\stdx{0.0}$  & $\stdx{1.1}$  & $\stdx{0.1}$  & $\stdx{0.0}$  & $\stdx{0.3}$  & $\stdx{0.3}$  & $\stdx{3.0}$  & $\stdx{1.8}$  & $\stdx{3.1}$  & $\stdx{1.5}$  & $\stdx{0.2}$  & $\stdx{0.5}$ \\  [-0.3ex] 
 \hline
\rowcolor{gray!20} &   & mean & ${98.5}$  & ${87.1}$  & ${93.5}$  & ${65.8}$  & ${65.4}$  & ${68.5}$  & ${74.7}$  & ${81.9}$  & ${93.1}$  & ${64.9}$  & ${95.5}$  & ${84.2}$  & ${66.3}$  & ${95.7}$  & ${84.4}$  & ${91.5}$  & ${84.1}$  & ${72.8}$  & ${79.7}$  & ${81.5}$ \\ 
 \rowcolor{gray!20}\multirow{-2}{*}{SegFormer}& \multirow{-2}{*}{Cityscapes (\bbest{reference})}& std & $\stdx{0.0}$  & $\stdx{0.1}$  & $\stdx{0.2}$  & $\stdx{2.6}$  & $\stdx{1.1}$  & $\stdx{0.2}$  & $\stdx{0.3}$  & $\stdx{0.3}$  & $\stdx{0.1}$  & $\stdx{1.3}$  & $\stdx{0.1}$  & $\stdx{0.1}$  & $\stdx{0.6}$  & $\stdx{0.2}$  & $\stdx{2.5}$  & $\stdx{0.8}$  & $\stdx{1.9}$  & $\stdx{1.0}$  & $\stdx{0.1}$  & $\stdx{0.1}$ \\[-0.3ex] 

\specialrule{.14em}{.2em}{.2em}
 &&&&&&&&&&&&&&&&&&&&&&\\[-3ex]

\multirow{2}{*}{DeepLabV3+} & \multirow{2}{*}{GTAV}  & mean & $68.7$  & $26.3$  & $71.9$  & $\bbest{29.6}$  & $23.9$  & $33.6$  & $41.4$  & $17.2$  & $83.9$  & ${31.2}$  & $50.3$  & $64.7$  & $25.3$  & $64.8$  & ${30.2}$  & $18.0$  & $0.3$  & $20.3$  & $20.2$  & $38.0$ \\ 
& & std & $\stdx{7.7}$  & $ \stdx{2.7}$  & $\stdx{2.5}$  & $\stdx{2.7}$  & $\stdx{0.5}$  & $\stdx{1.1}$  & $\stdx{0.6}$  & $\stdx{1.4}$  & $\stdx{0.3}$  & $\stdx{2.3}$  & $\stdx{4.9}$  & $\stdx{0.9}$  & $\stdx{3.0}$  & $\stdx{15.7}$  & $\stdx{2.4}$  & $\stdx{3.9}$  & $\stdx{0.1}$  & $\stdx{1.6}$  & $\stdx{5.3}$  & $\stdx{0.6}$ \\ 
\multirow{2}{*}{DeepLabV3+} & \multirow{2}{*}{Synscapes}  & mean & $85.0$  & $\bbest{42.4}$  & $65.6$  & $23.7$  & $16.7$  & ${47.9}$  & ${52.1}$  & ${56.8}$  & $84.4$  & $28.9$  & $77.2$  & $65.7$  & $28.8$  & $\bbest{86.3}$  & $21.2$  & $10.7$  & $4.0$  & ${37.0}$  & $61.0$  & $47.1$ \\ 
& & std & $\stdx{2.1}$  & $\stdx{1.5}$  & $\stdx{3.5}$  & $\stdx{2.5}$  & $\stdx{3.5}$  & $\stdx{0.7}$  & $\stdx{1.1}$  & $\stdx{1.6}$  & $\stdx{0.2}$  & $\stdx{2.0}$  & $\stdx{2.4}$  & $\stdx{0.7}$  & $\stdx{1.3}$  & $\stdx{1.6}$  & $\stdx{2.5}$  & $\stdx{1.4}$  & $\stdx{0.5}$  & $\stdx{2.3}$  & $\stdx{0.4}$  & $\stdx{1.2}$ \\ 
\multirow{2}{*}{DeepLabV3+} & \multirow{2}{*}{{\thedataset}}   & mean & $\bbest{86.4}$  & $\bbest{42.7}$  & $\bbest{84.4}$  & $25.4$  & $\bbest{40.4}$  & $\bbest{49.6}$  & $\bbest{54.4}$  & $\bbest{62.4}$  & $\bbest{86.4}$  & $\bbest{32.3}$  & $\bbest{84.7}$  & $\bbest{72.5}$  & $\bbest{51.0}$  & $76.1$  & $\bbest{46.6}$  & $\bbest{54.6}$  & $\bbest{23.2}$  & $\bbest{47.2}$  & $\bbest{69.8}$  & $\bbest{57.4}$ \\  
 & & std & $\stdx{0.6}$  & $\stdx{1.1}$  & $\stdx{0.6}$  & $\stdx{3.8}$  & $\stdx{1.5}$  & $\stdx{0.6}$  & $\stdx{0.6}$  & $\stdx{0.0}$  & $\stdx{0.1}$  & $\stdx{1.4}$  & $\stdx{0.7}$  & $\stdx{0.5}$  & $\stdx{1.1}$  & $\stdx{3.8}$  & $\stdx{3.7}$  & $\stdx{10.9}$  & $\stdx{5.8}$  & $\stdx{1.3}$  & $\stdx{0.3}$  & $\stdx{1.3}$ \\  \hline
 
\multirow{2}{*}{DeepLabV3+} & \multirow{2}{*}{GTAV+Synscapes}  & mean & $88.8$  & $48.5$  & $85.5$  & $42.7$  & $45.1$  & $51.8$  & $59.4$  & $50.7$  & $\bbest{88.7}$  & $35.8$  & $76.6$  & $71.1$  & $30.1$  & ${90.3}$  & $49.1$  & $67.2$  & $43.6$  & $50.3$  & $57.1$  & $59.6$ \\ 
 & & std & $\stdx{1.4}$  & $\stdx{0.8}$  & $\stdx{0.9}$  & $\stdx{2.9}$  & $\stdx{1.5}$  & $\stdx{0.9}$  & $\stdx{1.2}$  & $\stdx{1.5}$  & $\stdx{0.3}$  & $\stdx{3.3}$  & $\stdx{2.2}$  & $\stdx{0.8}$  & $\stdx{2.1}$  & $\stdx{1.0}$  & $\stdx{2.5}$  & $\stdx{3.5}$  & $\stdx{3.7}$  & $\stdx{2.7}$  & $\stdx{1.8}$  & $\stdx{0.6}$ \\ 
 
\multirow{2}{*}{DeepLabV3+} & \multirow{2}{*}{GTAV+Synscapes+{\thedataset}}  & mean & $\bbest{90.7}$  & $\bbest{52.3}$  & $\bbest{88.9}$  & $\bbest{48.5}$  & $\bbest{50.6}$  & $\bbest{57.6}$  & $\bbest{63.6}$  & $\bbest{64.7}$  & $\bbest{89.6}$  & $\bbest{44.6}$  & $\bbest{91.6}$  & $\bbest{77.5}$  & $\bbest{54.2}$  & $\bbest{91.8}$  & $\bbest{64.7}$  & $\bbest{75.5}$  & $\bbest{56.5}$  & $\bbest{55.6}$  & $\bbest{71.2}$  & $\bbest{67.9}$ \\  
& & std & $\stdx{0.0}$  & $\stdx{0.5}$  & $\stdx{0.3}$  & $\stdx{1.6}$  & $\stdx{0.8}$  & $\stdx{0.1}$  & $\stdx{0.3}$  & $\stdx{1.3}$  & $\stdx{0.1}$  & $\stdx{0.8}$  & $\stdx{0.2}$  & $\stdx{0.4}$  & $\stdx{0.8}$  & $\stdx{0.4}$  & $\stdx{4.2}$  & $\stdx{3.6}$  & $\stdx{4.8}$  & $\stdx{0.5}$  & $\stdx{0.7}$  & $\stdx{0.7}$ \\ \hline

\multirow{2}{*}{Co-training} & \multirow{2}{*}{GTAV}    & mean & $89.9$  & $50.8$  & $\bbest{89.2}$  & ${39.6}$  & $36.1$  & $52.3$  & $57.0$  & $51.6$  & $89.5$  & $48.2$  & $90.0$  & $72.0$  & $46.9$  & $91.5$  & ${57.2}$  & $57.9$  & $0.0$  & $53.6$  & $64.7$  & $59.9$ \\ 
\UDAbackbone{(DeepLabV3+)} & & std & $\stdx{0.1}$  & $\stdx{0.4}$  & $\stdx{0.0}$  & $\stdx{0.9}$  & $\stdx{0.6}$  & $\stdx{0.1}$  & $\stdx{0.3}$  & $\stdx{0.5}$  & $\stdx{0.1}$  & $\stdx{0.6}$  & $\stdx{0.4}$  & $\stdx{0.4}$  & $\stdx{0.6}$  & $\stdx{0.1}$  & $\stdx{1.9}$  & $\stdx{3.0}$  & $\stdx{0.0}$  & $\stdx{0.8}$  & $\stdx{0.1}$  & $\stdx{0.3}$ \\
\multirow{2}{*}{Co-training} & \multirow{2}{*}{Synscapes}    & mean & $\bbest{91.3}$  & $\bbest{54.6}$  & $82.4$  & $36.7$  & $43.6$  & $\bbest{57.3}$  & $\bbest{64.8}$  & ${69.5}$  & $\gbbest{91.3}$  & $\bbest{51.1}$  & $\gbbest{94.1}$  & $77.3$  & $41.4$  & $\bbest{93.5}$  & $17.2$  & $19.6$  & $1.7$  & $\bbest{60.8}$  & $\gbbest{74.8}$  & $59.1$ \\ 
\UDAbackbone{(DeepLabV3+)} & & std & $\stdx{0.1}$  & $\stdx{0.5}$  & $\stdx{0.1}$  & $\stdx{0.3}$  & $\stdx{0.2}$  & $\stdx{0.3}$  & $\stdx{0.2}$  & $\stdx{0.3}$  & $\stdx{0.1}$  & $\stdx{0.4}$  & $\stdx{0.1}$  & $\stdx{0.2}$  & $\stdx{1.6}$  & $\stdx{0.1}$  & $\stdx{1.1}$  & $\stdx{0.2}$  & $\stdx{0.1}$  & $\stdx{0.9}$  & $\stdx{0.3}$  & $\stdx{0.1}$ \\ 
\multirow{2}{*}{Co-training} & \multirow{2}{*}{{\thedataset}}    & mean & $\gbbest{91.8}$  & $\bbest{53.9}$  & $\bbest{88.9}$  & $\bbest{40.9}$  & $\bbest{51.2}$  & $\bbest{56.7}$  & $59.4$  & $\bbest{71.3}$  & $90.1$  & $\bbest{51.5}$  & $85.2$  & $\bbest{80.7}$  & $\bbest{60.0}$  & $\gbbest{94.3}$  & $\gbbest{79.4}$  & $\bbest{83.7}$  & $\bbest{62.4}$  & $\bbest{61.7}$  & $72.8$  & $\bbest{70.3}$ \\  
\UDAbackbone{(DeepLabV3+)} & & std & $\stdx{0.1}$  & $\stdx{0.4}$  & $\stdx{0.1}$  & $\stdx{1.1}$  & $\stdx{0.2}$  & $\stdx{0.3}$  & $\stdx{0.3}$  & $\stdx{0.2}$  & $\stdx{0.0}$  & $\stdx{0.4}$  & $\stdx{0.2}$  & $\stdx{0.1}$  & $\stdx{0.5}$  & $\stdx{0.0}$  & $\stdx{1.2}$  & $\stdx{1.1}$  & $\stdx{2.1}$  & $\stdx{0.2}$  & $\stdx{0.1}$  & $\stdx{0.3}$ \\ \hline

\multirow{2}{*}{Co-training} & \multirow{2}{*}{GTAV+Synscapes+{\thedataset}}  & mean & $\gbbest{96.8}$  & $\gbbest{76.9}$  & $\gbbest{91.8}$  & $\gbbest{56.4}$  & $\gbbest{58.3}$  & $\gbbest{62.4}$  & $\gbbest{67.4}$  & $\gbbest{75.0}$  & $\gbbest{91.7}$  & $\gbbest{57.2}$  & $\gbbest{94.6}$  & $\gbbest{82.1}$  & $\gbbest{62.0}$  & $\gbbest{94.8}$  & $\gbbest{82.7}$  & $\gbbest{90.0}$  & $\gbbest{76.4}$  & $\gbbest{65.2}$  & $\gbbest{75.0}$  & $\gbbest{76.7}$ \\
\UDAbackbone{(DeepLabV3+)} & & std & $\stdx{0.1}$  & $\stdx{0.8}$  & $\stdx{0.1}$  & $\stdx{0.0}$  & $\stdx{0.5}$  & $\stdx{0.2}$  & $\stdx{0.4}$  & $\stdx{0.2}$  & $\stdx{0.0}$  & $\stdx{0.3}$  & $\stdx{0.1}$  & $\stdx{0.2}$  & $\stdx{1.1}$  & $\stdx{0.0}$  & $\stdx{1.2}$  & $\stdx{0.5}$  & $\stdx{0.6}$  & $\stdx{0.7}$  & $\stdx{0.2}$  & $\stdx{0.1}$ \\ \hline

\rowcolor{gray!20}  &   & mean & ${98.2}$  & ${85.8}$  & ${93.1}$  & ${58.6}$  & ${64.1}$  & ${66.6}$  & ${71.4}$  & ${80.1}$  & ${92.8}$  & ${66.1}$  & ${95.2}$  & ${83.0}$  & ${63.8}$  & ${95.5}$  & ${79.1}$  & ${89.1}$  & ${83.8}$  & ${70.5}$  & ${78.4}$  & ${79.8}$ \\ 
\rowcolor{gray!20} \multirow{-2}{*}{DeepLabV3+} & \multirow{-2}{*}{Cityscapes (\bbest{reference})} & std & $\stdx{0.1}$  & $\stdx{0.2}$  & $\stdx{0.1}$  & $\stdx{0.7}$  & $\stdx{0.9}$  & $\stdx{0.1}$  & $\stdx{0.0}$  & $\stdx{0.3}$  & $\stdx{0.1}$  & $\stdx{1.0}$  & $\stdx{0.0}$  & $\stdx{0.2}$  & $\stdx{0.5}$  & $\stdx{0.1}$  & $\stdx{1.1}$  & $\stdx{0.8}$  & $\stdx{1.2}$  & $\stdx{0.8}$  & $\stdx{0.1}$  & $\stdx{0.1}$ \\[-0.3ex] 
 
\noalign{\hrule height 2pt}

\end{tabular}
}
\end{table*}

\begin{table*}[t]
\vspace{-4mm}
\caption{Analogous to Table \ref{tab:cityscapes_quantitative} but for BDD100K as target. (*) The bad results on the class Train are due to inconsistencies in the validation set labeling as we will illustrate in Section \ref{ssec:errorshumanlabels}.}
\label{tab:bdd_quantitative}
\centering
\setlength{\tabcolsep}{0.9mm} 
\scalebox{0.8}[0.8]{
\begin{tabular}{l|c|c|ccccccccccccccccccc!{\vrule width 1pt}c!{\vrule width 1pt}}

Method & Source & & \rotatebox[origin=c]{90}{\footnotesize Road} & \rotatebox[origin=c]{90}{\footnotesize Sidewalk} & \rotatebox[origin=c]{90}{\footnotesize Building} & \rotatebox[origin=c]{90}{\footnotesize Wall} & \rotatebox[origin=c]{90}{\footnotesize Fence} & \rotatebox[origin=c]{90}{\footnotesize Pole} & \rotatebox[origin=c]{90}{\footnotesize Traffic Light} & \rotatebox[origin=c]{90}{\footnotesize Traffic Sign} & \rotatebox[origin=c]{90}{\footnotesize Vegetation} & \rotatebox[origin=c]{90}{\footnotesize Terrain} & \rotatebox[origin=c]{90}{\footnotesize Sky} & \rotatebox[origin=c]{90}{\footnotesize Person} & \rotatebox[origin=c]{90}{\footnotesize Rider} & \rotatebox[origin=c]{90}{\footnotesize Car} & \rotatebox[origin=c]{90}{\footnotesize Truck} & \rotatebox[origin=c]{90}{\footnotesize Bus} & \rotatebox[origin=c]{90}{{\footnotesize Train}*} & \rotatebox[origin=c]{90}{\footnotesize Motorbike} & \rotatebox[origin=c]{90}{\footnotesize Bike} & mIoU  \\

\noalign{\hrule height 2pt}

\multirow{2}{*}{SegFormer} & \multirow{2}{*}{GTAV}  & mean & $76.2$  & $25.9$  & $\bbest{75.8}$  & $\bbest{8.8}$  & $\bbest{35.0}$  & $39.2$  & $\bbest{42.0}$  & $24.8$  & $72.2$  & $\bbest{35.1}$  & $\bbest{88.3}$  & $\bbest{55.6}$  & $\bbest{26.2}$  & $74.8$  & $\bbest{26.7}$  & $13.2$  & $\gbbest{0.0}$  & $\bbest{41.0}$  & $21.9$  & $\bbest{41.2}$ \\ 
& & std & $\stdx{2.9}$  & $\stdx{0.6}$  & $\stdx{0.4}$  & $\stdx{3.1}$  & $\stdx{1.7}$  & $\stdx{0.2}$  & $\stdx{1.2}$  & $\stdx{1.5}$  & $\stdx{1.3}$  & $\stdx{0.7}$  & $\stdx{0.6}$  & $\stdx{0.6}$  & $\stdx{1.6}$  & $\stdx{5.5}$  & $\stdx{1.3}$  & $\stdx{2.6}$  & $\stdx{0.0}$  & $\stdx{1.8}$  & $\stdx{6.7}$  & $\stdx{0.2}$ \\ 

\multirow{2}{*}{SegFormer} & \multirow{2}{*}{Synscapes}  & mean & $79.8$  & $\bbest{29.9}$  & $62.6$  & $2.9$  & $15.8$  & $29.9$  & $27.9$  & ${32.9}$  & $\bbest{73.6}$  & $20.2$  & $86.1$  & $49.4$  & $18.2$  & $72.8$  & $10.3$  & $13.7$  & $\gbbest{0.0}$  & $38.9$  & $\bbest{41.3}$  & $37.2$ \\ 
& & std & $\stdx{2.6}$  & $\stdx{0.6}$  & $\stdx{1.2}$  & $\stdx{0.1}$  & $\stdx{2.7}$  & $\stdx{1.1}$  & $\stdx{0.8}$  & $\stdx{1.1}$  & $\stdx{0.8}$  & $\stdx{2.3}$  & $\stdx{0.8}$  & $\stdx{1.3}$  & $\stdx{1.9}$  & $\stdx{2.3}$  & $\stdx{1.5}$  & $\stdx{3.3}$  & $\stdx{0.0}$  & $\stdx{3.3}$  & $\stdx{3.6}$  & $\stdx{0.3}$ \\ 

\multirow{2}{*}{SegFormer} & \multirow{2}{*}{{\thedataset}}   & mean & $\bbest{86.3}$  & $24.1$  & $63.2$  & $4.6$  & $25.4$  & $\bbest{40.6}$  & $39.8$  & $\bbest{40.1}$  & $65.6$  & $19.7$  & $78.6$  & $45.7$  & $22.6$  & $\bbest{82.7}$  & $21.7$  & $\bbest{51.3}$  & $\gbbest{0.0}$  & $38.6$  & $34.3$  & $\bbest{41.3}$ \\ 
& & std & $\stdx{0.8}$  & $\stdx{3.9}$  & $\stdx{2.3}$  & $\stdx{0.4}$  & $\stdx{5.9}$  & $\stdx{0.3}$  & $\stdx{1.2}$  & $\stdx{1.5}$  & $\stdx{2.2}$  & $\stdx{2.4}$  & $\stdx{1.7}$  & $\stdx{2.6}$  & $\stdx{3.8}$  & $\stdx{0.1}$  & $\stdx{2.2}$  & $\stdx{4.2}$  & $\stdx{0.1}$  & $\stdx{1.6}$  & $\stdx{4.2}$  & $\stdx{0.8}$ \\  \hline

\multirow{2}{*}{SegFormer} & \multirow{2}{*}{GTAV+Synscapes}  & mean & $82.4$  & $31.6$  & ${78.7}$  & $\bbest{18.4}$  & $\bbest{38.0}$  & $43.0$  & ${45.1}$  & $30.1$  & $\bbest{76.8}$  & $\bbest{35.8}$  & $\bbest{89.4}$  & $\bbest{61.4}$  & $24.2$  & $76.2$  & $30.5$  & $\bbest{44.9}$  & $\gbbest{0.0}$  & $52.4$  & $50.1$  & $47.8$ \\ 
& & std & $\stdx{3.8}$  & $\stdx{1.2}$  & $\stdx{1.0}$  & $\stdx{1.0}$  & $\stdx{0.8}$  & $\stdx{0.9}$  & $\stdx{1.0}$  & $\stdx{1.6}$  & $\stdx{0.6}$  & $\stdx{0.5}$  & $\stdx{0.6}$  & $\stdx{0.6}$  & $\stdx{2.1}$  & $\stdx{4.0}$  & $\stdx{1.4}$  & $\stdx{11.7}$  & $\stdx{0.0}$  & $\stdx{1.5}$  & $\stdx{3.1}$  & $\stdx{1.2}$ \\

\multirow{2}{*}{SegFormer} & \multirow{2}{*}{GTAV+Synscapes+{\thedataset}}   & mean & $\bbest{87.7}$  & $\bbest{37.8}$  & $\bbest{80.1}$  & $\bbest{18.1}$  & $37.0$  & $\bbest{45.5}$  & $\bbest{46.8}$  & $\bbest{42.1}$  & $75.1$  & $\bbest{36.5}$  & $\bbest{89.2}$  & $\bbest{62.0}$  & $\bbest{41.6}$  & $\bbest{82.3}$  & $\bbest{35.0}$  & $\bbest{45.8}$  & $\gbbest{0.0}$  & $\bbest{56.8}$  & $\bbest{59.2}$  & $\bbest{51.5}$ \\ 
& & std & $\stdx{1.4}$  & $\stdx{2.6}$  & $\stdx{0.1}$  & $\stdx{4.1}$  & $\stdx{1.8}$  & $\stdx{0.2}$  & $\stdx{0.6}$  & $\stdx{1.0}$  & $\stdx{1.0}$  & $\stdx{1.6}$  & $\stdx{0.7}$  & $\stdx{1.5}$  & $\stdx{5.3}$  & $\stdx{2.1}$  & $\stdx{1.6}$  & $\stdx{2.9}$  & $\stdx{0.0}$  & $\stdx{0.6}$  & $\stdx{1.2}$  & $\stdx{0.5}$ \\   \hline

 \multirow{2}{*}{HRDA} & \multirow{2}{*}{GTAV}   & mean & $90.0$  & $28.2$  & $\bbest{81.3}$  & $\gbbest{29.9}$  & ${38.6}$  & ${42.6}$  & $\bbest{48.2}$  & $38.6$  & ${80.6}$  & ${41.3}$  & $\bbest{91.3}$  & ${60.5}$  & $\bbest{38.5}$  & $85.2$  & $42.1$  & $65.0$  & $\gbbest{0.0}$  & $43.1$  & $34.1$  & $\bbest{51.5}$ \\  
\UDAbackbone{(DAFormer*)} & & std & $\stdx{0.9}$  & $\stdx{10.8}$  & $\stdx{0.5}$  & $\stdx{2.0}$  & $\stdx{1.0}$  & $\stdx{0.4}$  & $\stdx{0.1}$  & $\stdx{0.4}$  & $\stdx{0.3}$  & $\stdx{2.2}$  & $\stdx{0.2}$  & $\stdx{0.6}$  & $\stdx{3.8}$  & $\stdx{0.5}$  & $\stdx{2.1}$  & $\stdx{3.2}$  & $\stdx{0.0}$  & $\stdx{4.1}$  & $\stdx{2.7}$  & $\stdx{1.2}$ \\ 

\multirow{2}{*}{HRDA} & \multirow{2}{*}{Synscapes}  & mean & $61.5$  & $18.8$  & $78.1$  & $9.0$  & $15.9$  & $38.6$  & $43.4$  & $26.3$  & $79.4$  & $16.3$  & $\bbest{91.3}$  & $58.2$  & $27.5$  & $85.1$  & $25.7$  & $49.1$  & $\gbbest{0.0}$  & $\bbest{46.4}$  & ${51.4}$  & $43.3$ \\ 
\UDAbackbone{(DAFormer*)} & & std & $\stdx{24.6}$  & $\stdx{9.2}$  & $\stdx{1.0}$  & $\stdx{4.6}$  & $\stdx{2.6}$  & $\stdx{1.0}$  & $\stdx{1.4}$  & $\stdx{1.7}$  & $\stdx{0.7}$  & $\stdx{10.6}$  & $\stdx{0.5}$  & $\stdx{2.1}$  & $\stdx{0.9}$  & $\stdx{1.6}$  & $\stdx{9.6}$  & $\stdx{6.4}$  & $\stdx{0.0}$  & $\stdx{2.4}$  & $\stdx{0.6}$  & $\stdx{2.9}$ \\ 

\multirow{2}{*}{HRDA} & \multirow{2}{*}{{\thedataset}}   & mean & $\gbbest{92.1}$  & $\bbest{45.9}$  & $\bbest{81.0}$  & $8.3$  & $\bbest{40.2}$  & $\bbest{43.7}$  & $\bbest{49.1}$  & $\gbbest{53.6}$  & $\bbest{81.7}$  & $\gbbest{42.9}$  & $\bbest{91.4}$  & $\bbest{64.5}$  & $34.4$  & $\gbbest{87.9}$  & $\bbest{45.0}$  & $\bbest{69.2}$  & $\gbbest{0.0}$  & $44.0$  & $\bbest{54.1}$  & $\bbest{54.2}$ \\ 
\UDAbackbone{(DAFormer*)} & & std & $\stdx{0.4}$  & $\stdx{1.1}$  & $\stdx{0.4}$  & $\stdx{3.1}$  & $\stdx{1.3}$  & $\stdx{0.8}$  & $\stdx{0.6}$  & $\stdx{0.5}$  & $\stdx{0.4}$  & $\stdx{2.2}$  & $\stdx{0.3}$  & $\stdx{2.4}$  & $\stdx{7.9}$  & $\stdx{0.3}$  & $\stdx{1.4}$  & $\stdx{1.4}$  & $\stdx{0.0}$  & $\stdx{10.2}$  & $\stdx{0.5}$  & $\stdx{0.8}$ \\    \hline

\multirow{2}{*}{HRDA} & \multirow{2}{*}{GTAV+Synscapes+{\thedataset}}  & mean & $90.5$  & $\gbbest{56.7}$  & $\gbbest{82.6}$  & $\gbbest{29.3}$  & $\gbbest{41.8}$  & $\gbbest{49.5}$  & $\gbbest{52.5}$  & $52.4$  & $\gbbest{82.7}$  & $\gbbest{43.8}$  & $\gbbest{92.7}$  & $\gbbest{67.0}$  & $\gbbest{47.5}$  & $81.9$  & $\gbbest{49.8}$  & $\gbbest{76.1}$  & $\gbbest{0.0}$  & $\gbbest{55.1}$  & $\gbbest{62.0}$  & $\gbbest{58.6}$ \\ 
\UDAbackbone{(DAFormer*)} & & std & $\stdx{0.5}$  & $\stdx{0.9}$  & $\stdx{0.1}$  & $\stdx{1.1}$  & $\stdx{1.8}$  & $\stdx{0.1}$  & $\stdx{1.1}$  & $\stdx{0.9}$  & $\stdx{0.3}$  & $\stdx{0.6}$  & $\stdx{0.2}$  & $\stdx{1.3}$  & $\stdx{2.5}$  & $\stdx{1.5}$  & $\stdx{1.0}$  & $\stdx{3.1}$  & $\stdx{0.0}$  & $\stdx{1.7}$  & $\stdx{1.0}$  & $\stdx{0.4}$ \\  [-0.35ex] 

\hline

\rowcolor{gray!20}  &  & mean & $95.0$  & $66.1$  & $87.2$  & $34.2$  & $52.7$  & $51.0$  & $56.8$  & $57.4$  & $87.4$  & $53.0$  & $95.8$  & $66.4$  & $29.7$  & $90.7$  & $61.3$  & $81.8$  & $0.0$  & $48.7$  & $52.8$  & $61.5$ \\
\rowcolor{gray!20} \multirow{-2}{*}{SegFormer} & \multirow{-2}{*}{BDD100K (\bbest{reference})} & std & $\stdx{0.1}$  & $\stdx{0.6}$  & $\stdx{0.0}$  & $\stdx{0.2}$  & $\stdx{0.1}$  & $\stdx{0.4}$  & $\stdx{0.5}$  & $\stdx{0.3}$  & $\stdx{0.1}$  & $\stdx{1.0}$  & $\stdx{0.2}$  & $\stdx{0.6}$  & $\stdx{12.7}$  & $\stdx{0.5}$  & $\stdx{1.1}$  & $\stdx{2.0}$  & $\stdx{0.0}$  & $\stdx{7.9}$  & $\stdx{3.8}$  & $\stdx{1.2}$ \\ [-0.35ex] 

\specialrule{.14em}{.2em}{.2em}
 &&&&&&&&&&&&&&&&&&&&&&\\[-3ex]

\multirow{2}{*}{DeepLabV3+} & \multirow{2}{*}{GTAV} & mean & $63.6$  & ${25.9}$  & $59.4$  & $\bbest{6.3}$  & $\bbest{27.6}$  & $32.3$  & $\bbest{36.0}$  & $21.3$  & $66.2$  & $21.1$  & $77.1$  & $\bbest{52.2}$  & $16.9$  & $\bbest{68.2}$  & $18.4$  & $17.9$  & $\gbbest{0.0}$  & $29.0$  & $6.1$  & $34.0$ \\ 
& & std & $\stdx{3.7}$  & $\stdx{1.8}$  & $\stdx{2.5}$  & $\stdx{0.5}$  & $\stdx{2.1}$  & $\stdx{0.8}$  & $\stdx{0.5}$  & $\stdx{1.7}$  & $\stdx{0.6}$  & $\stdx{3.0}$  & $\stdx{2.6}$  & $\stdx{1.3}$  & $\stdx{1.9}$  & $\stdx{1.2}$  & $\stdx{1.0}$  & $\stdx{1.5}$  & $\stdx{0.0}$  & $\stdx{7.3}$  & $\stdx{1.8}$  & $\stdx{0.1}$ \\ 

\multirow{2}{*}{DeepLabV3+} & \multirow{2}{*}{Synscapes}  & mean & $66.5$  & $21.1$  & $32.9$  & $1.1$  & $5.1$  & $22.3$  & $25.3$  & ${23.2}$  & $61.9$  & $6.5$  & $57.3$  & $47.3$  & ${17.9}$  & $63.3$  & $1.6$  & $5.7$  & $\gbbest{0.0}$  & $30.8$  & $34.0$  & $27.6$ \\ 
& & std & $\stdx{5.4}$  & $\stdx{2.5}$  & $\stdx{3.7}$  & $\stdx{0.5}$  & $\stdx{0.8}$  & $\stdx{2.8}$  & $\stdx{0.9}$  & $\stdx{1.5}$  & $\stdx{3.8}$  & $\stdx{0.4}$  & $\stdx{6.5}$  & $\stdx{3.5}$  & $\stdx{3.4}$  & $\stdx{2.3}$  & $\stdx{0.6}$  & $\stdx{0.5}$  & $\stdx{0.0}$  & $\stdx{2.5}$  & $\stdx{4.2}$  & $\stdx{1.1}$ \\ 

\multirow{2}{*}{DeepLabV3+} & \multirow{2}{*}{{\thedataset}}  & mean & $\bbest{80.1}$  & $\bbest{29.3}$  & $\bbest{64.8}$  & $3.3$  & $15.8$  & $\bbest{33.6}$  & $31.3$  & $\bbest{33.2}$  & $\bbest{74.1}$  & $\bbest{24.2}$  & $\bbest{79.5}$  & $42.5$  & $\bbest{22.4}$  & $\bbest{68.7}$  & $\bbest{19.6}$  & $\bbest{22.6}$  & $\gbbest{0.0}$  & $\bbest{45.3}$  & $\bbest{50.6}$  & $\bbest{39.0}$ \\
& & std & $\stdx{1.6}$  & $\stdx{1.4}$  & $\stdx{0.8}$  & $\stdx{0.2}$  & $\stdx{0.1}$  & $\stdx{0.3}$  & $\stdx{0.4}$  & $\stdx{0.6}$  & $\stdx{0.3}$  & $\stdx{1.6}$  & $\stdx{1.0}$  & $\stdx{7.8}$  & $\stdx{8.0}$  & $\stdx{1.4}$  & $\stdx{1.7}$  & $\stdx{5.2}$  & $\stdx{0.0}$  & $\stdx{0.6}$  & $\stdx{1.2}$  & $\stdx{0.5}$ \\   \hline

\multirow{2}{*}{DeepLabV3+} & \multirow{2}{*}{GTAV+Synscapes} & mean & $67.2$  & $30.2$  & $74.7$  & $12.9$  & $29.9$  & $\bbest{37.7}$  & $\bbest{39.6}$  & ${27.7}$  & $\bbest{75.0}$  & $25.1$  & $86.5$  & $\bbest{57.9}$  & $\bbest{37.1}$  & $62.0$  & $20.7$  & $25.0$  & $\gbbest{0.1}$  & $\bbest{47.5}$  & $50.5$  & $42.5$ \\ 
& & std & $\stdx{1.3}$  & $\stdx{2.3}$  & $\stdx{0.9}$  & $\stdx{3.0}$  & $\stdx{2.1}$  & $\stdx{2.0}$  & $\stdx{2.5}$  & $\stdx{0.9}$  & $\stdx{0.3}$  & $\stdx{1.6}$  & $\stdx{1.0}$  & $\stdx{1.6}$  & $\stdx{3.4}$  & $\stdx{1.6}$  & $\stdx{0.1}$  & $\stdx{1.6}$  & $\stdx{0.0}$  & $\stdx{1.8}$  & $\stdx{5.2}$  & $\stdx{1.1}$ \\ 

\multirow{2}{*}{DeepLabV3+} & \multirow{2}{*}{GTAV+Synscapes+{\thedataset}}   & mean & $\bbest{78.8}$  & $\bbest{34.7}$  & $\bbest{76.4}$  & $11.2$  & $\bbest{32.6}$  & $36.9$  & $37.0$  & $\bbest{29.5}$  & $\bbest{75.8}$  & $\bbest{33.1}$  & $\bbest{87.7}$  & $56.9$  & $26.5$  & $\bbest{71.2}$  & $\bbest{25.6}$  & $\bbest{45.1}$  & $\gbbest{0.0}$  & $\bbest{47.1}$  & $\bbest{55.8}$  & $\bbest{45.4}$ \\
& & std & $\stdx{0.2}$  & $\stdx{1.5}$  & $\stdx{0.5}$  & $\stdx{1.5}$  & $\stdx{2.1}$  & $\stdx{0.4}$  & $\stdx{0.3}$  & $\stdx{0.4}$  & $\stdx{0.7}$  & $\stdx{0.1}$  & $\stdx{0.5}$  & $\stdx{0.6}$  & $\stdx{2.9}$  & $\stdx{1.5}$  & $\stdx{0.3}$  & $\stdx{2.3}$  & $\stdx{0.0}$  & $\stdx{0.1}$  & $\stdx{1.4}$  & $\stdx{0.5}$ \\  \hline 

\multirow{2}{*}{Co-training} & \multirow{2}{*}{GTAV}    & mean & $82.7$  & $\gbbest{56.0}$  & $\bbest{77.9}$  & $\bbest{20.8}$  & $\bbest{41.7}$  & $36.9$  & $\bbest{34.9}$  & $19.7$  & $74.8$  & $\gbbest{37.4}$  & $87.5$  & $54.8$  & ${40.3}$  & $71.3$  & $27.5$  & $4.5$  & $\gbbest{0.4}$  & $38.4$  & $19.1$  & $43.5$ \\ 
\UDAbackbone{(DeepLabV3+)} & & std & $\stdx{0.1}$  & $\stdx{1.1}$  & $\stdx{0.3}$  & $\stdx{0.2}$  & $\stdx{1.4}$  & $\stdx{0.1}$  & $\stdx{0.8}$  & $\stdx{0.4}$  & $\stdx{0.3}$  & $\stdx{0.0}$  & $\stdx{0.2}$  & $\stdx{0.9}$  & $\stdx{0.5}$  & $\stdx{0.3}$  & $\stdx{0.6}$  & $\stdx{0.2}$  & $\stdx{0.4}$  & $\stdx{0.5}$  & $\stdx{3.6}$  & $\stdx{0.3}$ \\ 
\multirow{2}{*}{Co-training} & \multirow{2}{*}{Synscapes}   & mean & $\bbest{89.9}$  & $35.5$  & $52.7$  & $1.5$  & $6.9$  & $\bbest{39.5}$  & $32.2$  & $25.9$  & $\gbbest{78.6}$  & $25.5$  & $77.2$  & $\bbest{60.5}$  & $30.2$  & ${85.5}$  & $1.7$  & $29.6$  & $\gbbest{0.0}$  & $\gbbest{50.3}$  & $54.4$  & $40.9$ \\ 
\UDAbackbone{(DeepLabV3+)} & & std & $\stdx{0.1}$  & $\stdx{0.2}$  & $\stdx{0.4}$  & $\stdx{0.1}$  & $\stdx{0.3}$  & $\stdx{0.1}$  & $\stdx{0.2}$  & $\stdx{0.2}$  & $\stdx{0.2}$  & $\stdx{1.1}$  & $\stdx{0.6}$  & $\stdx{0.5}$  & $\stdx{4.5}$  & $\stdx{0.1}$  & $\stdx{0.3}$  & $\stdx{0.3}$  & $\stdx{0.0}$  & $\stdx{2.4}$  & $\stdx{0.5}$  & $\stdx{0.3}$ \\
\multirow{2}{*}{Co-training} & \multirow{2}{*}{{\thedataset}}  & mean & $\bbest{90.5}$  & $36.3$  & $76.9$  & $10.0$  & $29.2$  & $31.2$  & $30.1$  & $\bbest{32.3}$  & $74.8$  & $\gbbest{37.7}$  & $\gbbest{88.7}$  & $47.8$  & $\bbest{44.8}$  & $\gbbest{86.7}$  & $\bbest{37.7}$  & $\bbest{65.4}$  & $\gbbest{0.0}$  & $37.1$  & $43.1$  & $\bbest{47.4}$ \\  
\UDAbackbone{(DeepLabV3+)} & & std & $\stdx{0.0}$  & $\stdx{0.9}$  & $\stdx{0.3}$  & $\stdx{0.0}$  & $\stdx{0.8}$  & $\stdx{0.3}$  & $\stdx{0.8}$  & $\stdx{0.7}$  & $\stdx{0.2}$  & $\stdx{0.3}$  & $\stdx{0.1}$  & $\stdx{4.3}$  & $\stdx{1.9}$  & $\stdx{0.1}$  & $\stdx{1.5}$  & $\stdx{2.6}$  & $\stdx{0.0}$  & $\stdx{0.7}$  & $\stdx{1.7}$  & $\stdx{0.3}$ \\ \hline

\multirow{2}{*}{Co-training} & \multirow{2}{*}{GTAV+Synscapes+{\thedataset}}  & mean & $\gbbest{92.1}$  & $44.4$  & $\gbbest{81.1}$  & $\gbbest{24.2}$  & $\gbbest{44.1}$  & $\gbbest{43.2}$  & $\gbbest{46.6}$  & $\gbbest{40.3}$  & $72.4$  & $36.4$  & $87.0$  & $\gbbest{64.0}$  & $\gbbest{57.7}$  & $\gbbest{87.5}$  & $\gbbest{46.0}$  & $\gbbest{76.9}$  & $\gbbest{0.7}$  & $\gbbest{51.1}$  & $\gbbest{55.7}$  & $\gbbest{55.3}$ \\ 
\UDAbackbone{(DeepLabV3+)} & & std & $\stdx{0.1}$  & $\stdx{0.6}$  & $\stdx{0.1}$  & $\stdx{2.2}$  & $\stdx{0.6}$  & $\stdx{0.1}$  & $\stdx{0.2}$  & $\stdx{0.1}$  & $\stdx{0.2}$  & $\stdx{0.6}$  & $\stdx{0.1}$  & $\stdx{0.7}$  & $\stdx{1.4}$  & $\stdx{0.0}$  & $\stdx{0.7}$  & $\stdx{0.7}$  & $\stdx{0.2}$  & $\stdx{0.5}$  & $\stdx{0.9}$  & $\stdx{0.1}$ \\  \hline

\rowcolor{gray!20}  &  & mean & ${94.7}$  & ${65.7}$  & ${86.5}$  & ${36.1}$  & ${53.8}$  & ${50.9}$  & ${54.4}$  & ${55.4}$  & ${86.8}$  & ${50.7}$  & ${95.1}$  & ${65.0}$  & ${45.9}$  & ${90.6}$  & ${59.7}$  & ${81.0}$  & ${0.0}$  & ${47.6}$  & $43.4$  & ${61.2}$ \\ 
\rowcolor{gray!20} \multirow{-2}{*}{DeepLabV3+}& \multirow{-2}{*}{BDD100K (\bbest{reference)}}  & std & $\stdx{0.1}$  & $\stdx{0.1}$  & $\stdx{0.1}$  & $\stdx{0.9}$  & $\stdx{0.6}$  & $\stdx{0.3}$  & $\stdx{2.2}$  & $\stdx{0.2}$  & $\stdx{0.2}$  & $\stdx{0.5}$  & $\stdx{0.1}$  & $\stdx{0.6}$  & $\stdx{2.9}$  & $\stdx{0.1}$  & $\stdx{0.8}$  & $\stdx{1.9}$  & $\stdx{0.0}$  & $\stdx{3.0}$  & $\stdx{3.5}$  & $\stdx{0.3}$ \\[-0.3ex] 

\noalign{\hrule height 2pt}

\end{tabular}
}
\end{table*}

\begin{table*}[t]
\vspace{-4mm}
\caption{Analogous to Table \ref{tab:cityscapes_quantitative} but for Mapillary Vistas as target.}
\label{tab:mapillary_quantitative}
\centering
\setlength{\tabcolsep}{0.9mm} 
\scalebox{0.8}[0.8]{
\begin{tabular}{l|c|c|ccccccccccccccccccc!{\vrule width 1pt}c!{\vrule width 1pt}}

Method & Source & & \rotatebox[origin=c]{90}{\footnotesize Road} & \rotatebox[origin=c]{90}{\footnotesize Sidewalk} & \rotatebox[origin=c]{90}{\footnotesize Building} & \rotatebox[origin=c]{90}{\footnotesize Wall} & \rotatebox[origin=c]{90}{\footnotesize Fence} & \rotatebox[origin=c]{90}{\footnotesize Pole} & \rotatebox[origin=c]{90}{\footnotesize Traffic Light} & \rotatebox[origin=c]{90}{\footnotesize Traffic Sign} & \rotatebox[origin=c]{90}{\footnotesize Vegetation} & \rotatebox[origin=c]{90}{\footnotesize Terrain} & \rotatebox[origin=c]{90}{\footnotesize Sky} & \rotatebox[origin=c]{90}{\footnotesize Person} & \rotatebox[origin=c]{90}{\footnotesize Rider} & \rotatebox[origin=c]{90}{\footnotesize Car} & \rotatebox[origin=c]{90}{\footnotesize Truck} & \rotatebox[origin=c]{90}{\footnotesize Bus} & \rotatebox[origin=c]{90}{\footnotesize Train} & \rotatebox[origin=c]{90}{\footnotesize Motorbike} & \rotatebox[origin=c]{90}{\footnotesize Bike} & mIoU  \\ 

\noalign{\hrule height 2pt}

\multirow{2}{*}{SegFormer} & \multirow{2}{*}{GTAV}  & mean & $83.3$  & $\bbest{40.2}$  & $\bbest{83.9}$  & $\bbest{32.0}$  & $41.4$  & $43.4$  & $53.1$  & $53.4$  & $77.6$  & $43.1$  & $\gbbest{94.8}$  & $70.1$  & $31.9$  & $86.2$  & ${49.3}$  & $42.0$  & $\bbest{21.1}$  & $43.6$  & $22.3$  & $53.3$ \\  
& & std & $\stdx{1.9}$  & $\stdx{3.1}$  & $\stdx{0.1}$  & $\stdx{1.4}$  & $\stdx{0.8}$  & $\stdx{0.5}$  & $\stdx{1.1}$  & $\stdx{1.7}$  & $\stdx{1.3}$  & $\stdx{3.2}$  & $\stdx{0.1}$  & $\stdx{0.8}$  & $\stdx{1.6}$  & $\stdx{1.4}$  & $\stdx{2.0}$  & $\stdx{5.1}$  & $\stdx{4.5}$  & $\stdx{2.4}$  & $\stdx{0.4}$  & $\stdx{0.8}$ \\ 

\multirow{2}{*}{SegFormer} & \multirow{2}{*}{Synscapes}  & mean & $83.6$  & $39.0$  & $72.1$  & $27.0$  & $26.2$  & $39.1$  & $50.2$  & ${66.8}$  & $\bbest{81.4}$  & $\bbest{49.5}$  & $93.3$  & $64.6$  & $43.8$  & $83.5$  & $30.2$  & $8.6$  & $1.9$  & $49.3$  & ${59.2}$  & $51.0$ \\  
& & std & $\stdx{2.3}$  & $\stdx{3.7}$  & $\stdx{1.0}$  & $\stdx{2.6}$  & $\stdx{3.0}$  & $\stdx{1.8}$  & $\stdx{1.6}$  & $\stdx{0.7}$  & $\stdx{0.8}$  & $\stdx{1.3}$  & $\stdx{0.7}$  & $\stdx{2.0}$  & $\stdx{3.6}$  & $\stdx{2.8}$  & $\stdx{2.2}$  & $\stdx{3.4}$  & $\stdx{1.1}$  & $\stdx{1.7}$  & $\stdx{0.7}$  & $\stdx{0.8}$ \\

\multirow{2}{*}{SegFormer} & \multirow{2}{*}{{\thedataset}}   & mean & $\bbest{88.0}$  & $33.2$  & $73.6$  & $13.5$  & $\bbest{44.5}$  & $\bbest{49.6}$  & $\bbest{59.2}$  & $\bbest{69.2}$  & $80.2$  & $48.2$  & $91.2$  & $\bbest{74.3}$  & $\bbest{55.8}$  & $\bbest{89.3}$  & $\bbest{52.1}$  & $\bbest{52.5}$  & $\bbest{21.0}$  & $\bbest{58.0}$  & $\bbest{64.0}$  & $\bbest{58.8}$ \\   
& & std & $\stdx{0.8}$  & $\stdx{4.6}$  & $\stdx{3.0}$  & $\stdx{2.7}$  & $\stdx{1.4}$  & $\stdx{0.2}$  & $\stdx{0.4}$  & $\stdx{0.4}$  & $\stdx{0.3}$  & $\stdx{0.8}$  & $\stdx{1.8}$  & $\stdx{0.2}$  & $\stdx{4.2}$  & $\stdx{0.3}$  & $\stdx{2.9}$  & $\stdx{6.9}$  & $\stdx{2.5}$  & $\stdx{1.3}$  & $\stdx{1.5}$  & $\stdx{0.3}$ \\   \hline

\multirow{2}{*}{SegFormer} & \multirow{2}{*}{GTAV+Synscapes}  & mean & $86.6$  & $46.8$  & $85.7$  & $38.3$  & $46.0$  & $47.5$  & $58.5$  & $60.3$  & $\bbest{79.8}$  & $\bbest{48.2}$  & $\gbbest{95.5}$  & $75.4$  & $51.5$  & $89.2$  & $56.3$  & $56.4$  & $28.8$  & $55.5$  & $55.8$  & $61.2$ \\  
& & std & $\stdx{0.4}$  & $\stdx{1.7}$  & $\stdx{0.1}$  & $\stdx{2.3}$  & $\stdx{1.6}$  & $\stdx{0.6}$  & $\stdx{0.5}$  & $\stdx{1.3}$  & $\stdx{0.8}$  & $\stdx{1.0}$  & $\stdx{0.1}$  & $\stdx{0.4}$  & $\stdx{3.1}$  & $\stdx{0.2}$  & $\stdx{2.4}$  & $\stdx{3.2}$  & $\stdx{8.6}$  & $\stdx{1.8}$  & $\stdx{4.5}$  & $\stdx{0.5}$ \\  

\multirow{2}{*}{SegFormer} & \multirow{2}{*}{GTAV+Synscapes+{\thedataset}}  & mean & $\bbest{90.1}$  & $\bbest{53.3}$  & $\bbest{86.7}$  & $\bbest{42.2}$  & $\bbest{52.4}$  & $\bbest{51.8}$  & $\bbest{62.3}$  & $\bbest{68.8}$  & $\bbest{80.4}$  & $\bbest{48.3}$  & $\bbest{95.6}$  & $\bbest{77.5}$  & $\bbest{61.1}$  & $\bbest{91.8}$  & $\bbest{68.3}$  & $\bbest{69.6}$  & $\bbest{52.3}$  & $\bbest{59.8}$  & $\bbest{63.6}$  & $\gbbest{67.2}$ \\
& & std & $\stdx{0.8}$  & $\stdx{2.3}$  & $\stdx{0.2}$  & $\stdx{1.0}$  & $\stdx{1.6}$  & $\stdx{0.5}$  & $\stdx{0.3}$  & $\stdx{0.3}$  & $\stdx{0.2}$  & $\stdx{0.8}$  & $\stdx{0.1}$  & $\stdx{0.8}$  & $\stdx{0.2}$  & $\stdx{0.2}$  & $\stdx{2.2}$  & $\stdx{3.7}$  & $\stdx{5.1}$  & $\stdx{1.1}$  & $\stdx{0.6}$  & $\stdx{0.6}$ \\   \hline

\multirow{2}{*}{HRDA} & \multirow{2}{*}{GTAV}   & mean & $\bbest{85.5}$  & $\bbest{21.0}$  & $\bbest{86.6}$  & $\bbest{46.0}$  & $35.2$  & $47.3$  & $58.0$  & $62.4$  & $66.0$  & $36.5$  & $92.3$  & $65.0$  & $21.7$  & $89.1$  & $58.7$  & $64.2$  & $5.6$  & $14.8$  & $0.0$  & $50.3$ \\
\UDAbackbone{(DAFormer*)} & & std & $\stdx{0.8}$  & $\stdx{7.0}$  & $\stdx{1.1}$  & $\stdx{0.9}$  & $\stdx{0.9}$  & $\stdx{0.4}$  & $\stdx{1.6}$  & $\stdx{8.1}$  & $\stdx{8.0}$  & $\stdx{13.6}$  & $\stdx{0.1}$  & $\stdx{9.7}$  & $\stdx{18.6}$  & $\stdx{0.8}$  & $\stdx{1.1}$  & $\stdx{5.7}$  & $\stdx{3.9}$  & $\stdx{14.8}$  & $\stdx{0.0}$  & $\stdx{0.3}$ \\ 

\multirow{2}{*}{HRDA} & \multirow{2}{*}{Synscapes}  & mean & $\bbest{84.6}$  & $12.8$  & $74.6$  & $26.1$  & $\bbest{43.4}$  & $46.8$  & $51.0$  & ${65.5}$  & $\gbbest{87.8}$  & $\bbest{42.2}$  & $\bbest{92.8}$  & $68.6$  & $51.3$  & $89.2$  & $56.7$  & $58.0$  & $9.8$  & $55.5$  & $62.3$  & $\bbest{56.8}$ \\
\UDAbackbone{(DAFormer*)} & & std & $\stdx{3.1}$  & $\stdx{12.3}$  & $\stdx{10.1}$  & $\stdx{1.4}$  & $\stdx{3.5}$  & $\stdx{2.7}$  & $\stdx{5.9}$  & $\stdx{1.9}$  & $\stdx{0.7}$  & $\stdx{27.6}$  & $\stdx{6.7}$  & $\stdx{6.2}$  & $\stdx{5.2}$  & $\stdx{1.1}$  & $\stdx{6.6}$  & $\stdx{5.7}$  & $\stdx{5.1}$  & $\stdx{4.1}$  & $\stdx{4.6}$  & $\stdx{5.1}$ \\  

\multirow{2}{*}{HRDA} & \multirow{2}{*}{{\thedataset}} & mean & $80.1$  & $7.2$  & $85.2$  & $7.9$  & $35.0$  & $\bbest{50.0}$  & $\bbest{63.4}$  & $\bbest{74.1}$  & $74.2$  & $0.0$  & $\bbest{93.4}$  & $\bbest{78.4}$  & $\bbest{57.0}$  & $\bbest{90.8}$  & $\bbest{66.4}$  & $\bbest{69.8}$  & $\bbest{29.0}$  & $\bbest{57.8}$  & $\bbest{67.7}$  & $\bbest{57.2}$ \\  
\UDAbackbone{(DAFormer*)} & & std & $\stdx{3.4}$  & $\stdx{9.8}$  & $\stdx{0.3}$  & $\stdx{9.4}$  & $\stdx{10.0}$  & $\stdx{0.1}$  & $\stdx{0.3}$  & $\stdx{0.7}$  & $\stdx{1.7}$  & $\stdx{0.0}$  & $\stdx{0.3}$  & $\stdx{0.2}$  & $\stdx{1.0}$  & $\stdx{0.5}$  & $\stdx{1.4}$  & $\stdx{4.9}$  & $\stdx{16.5}$  & $\stdx{2.0}$  & $\stdx{0.8}$  & $\stdx{1.9}$ \\  \hline

\multirow{2}{*}{HRDA} & \multirow{2}{*}{GTAV+Synscapes+{\thedataset}}  & mean & $88.3$  & $\gbbest{55.4}$  & $\gbbest{88.4}$  & $\gbbest{49.1}$  & $49.7$  & $\gbbest{52.2}$  & $\gbbest{63.9}$  & $\gbbest{75.9}$  & $\gbbest{86.9}$  & $\gbbest{61.1}$  & $\gbbest{97.2}$  & $\gbbest{78.7}$  & $\gbbest{60.9}$  & $\gbbest{91.7}$  & $\gbbest{71.3}$  & $52.7$  & $44.4$  & $\gbbest{58.9}$  & $65.4$  & $\gbbest{68.0}$ \\  
\UDAbackbone{(DAFormer*)} & & std & $\stdx{4.7}$  & $\stdx{5.1}$  & $\stdx{0.2}$  & $\stdx{1.2}$  & $\stdx{1.2}$  & $\stdx{0.5}$  & $\stdx{0.3}$  & $\stdx{0.2}$  & $\stdx{0.6}$  & $\stdx{0.2}$  & $\stdx{0.3}$  & $\stdx{0.3}$  & $\stdx{2.0}$  & $\stdx{0.1}$  & $\stdx{1.0}$  & $\stdx{31.1}$  & $\stdx{6.0}$  & $\stdx{1.6}$  & $\stdx{0.6}$  & $\stdx{2.2}$ \\ 
 \hline
\rowcolor{gray!20}&  & mean & $96.8$  & $82.1$  & $92.2$  & $60.9$  & $71.0$  & $67.1$  & $73.0$  & $83.3$  & $92.5$  & $75.6$  & $98.7$  & $81.0$  & $64.9$  & $93.9$  & $80.0$  & $87.6$  & $68.2$  & $67.8$  & $73.5$  & $79.5$ \\
\rowcolor{gray!20}  \multirow{-2}{*}{SegFormer} & \multirow{-2}{*}{Mapillary Vistas (\bbest{reference})} & std & $\stdx{0.1}$  & $\stdx{0.1}$  & $\stdx{0.3}$  & $\stdx{4.7}$  & $\stdx{1.4}$  & $\stdx{0.8}$  & $\stdx{0.5}$  & $\stdx{0.1}$  & $\stdx{0.0}$  & $\stdx{0.3}$  & $\stdx{0.0}$  & $\stdx{0.1}$  & $\stdx{0.1}$  & $\stdx{0.1}$  & $\stdx{0.6}$  & $\stdx{0.6}$  & $\stdx{4.5}$  & $\stdx{0.4}$  & $\stdx{0.2}$  & $\stdx{0.7}$ \\  [-0.3ex] 

\specialrule{.14em}{.2em}{.2em}
 &&&&&&&&&&&&&&&&&&&&&&\\[-3ex]

\multirow{2}{*}{DeeplabV3+} & \multirow{2}{*}{GTAV}  & mean & $68.4$  & $29.5$  & $55.6$  & $\bbest{18.9}$  & $32.2$  & $34.9$  & $\bbest{46.3}$  & $46.4$  & $75.8$  & $34.3$  & $75.2$  & $\bbest{68.1}$  & $30.8$  & $\bbest{81.0}$  & ${38.7}$  & $12.7$  & ${11.4}$  & $32.1$  & $23.6$  & $43.0$ \\ 
& & std & $\stdx{2.8}$  & $\stdx{1.9}$  & $\stdx{3.7}$  & $\stdx{2.8}$  & $\stdx{0.4}$  & $\stdx{1.9}$  & $\stdx{0.9}$  & $\stdx{1.5}$  & $\stdx{0.6}$  & $\stdx{0.8}$  & $\stdx{3.4}$  & $\stdx{1.0}$  & $\stdx{0.6}$  & $\stdx{1.6}$  & $\stdx{0.5}$  & $\stdx{1.3}$  & $\stdx{3.5}$  & $\stdx{1.0}$  & $\stdx{2.0}$  & $\stdx{1.1}$ \\  

\multirow{2}{*}{DeeplabV3+} & \multirow{2}{*}{Synscapes}   & mean & $67.6$  & $\bbest{30.7}$  & $45.0$  & $10.8$  & $10.2$  & $34.2$  & $38.7$  & $\bbest{56.1}$  & $73.4$  & $\bbest{39.1}$  & $78.7$  & $57.6$  & $35.8$  & $76.8$  & $7.3$  & $4.3$  & $0.2$  & $39.6$  & $\bbest{52.4}$  & $39.9$ \\ 
& & std & $\stdx{1.2}$  & $\stdx{0.8}$  & $\stdx{2.6}$  & $\stdx{0.7}$  & $\stdx{1.1}$  & $\stdx{0.7}$  & $\stdx{1.0}$  & $\stdx{1.2}$  & $\stdx{1.3}$  & $\stdx{0.9}$  & $\stdx{1.4}$  & $\stdx{0.5}$  & $\stdx{1.0}$  & $\stdx{2.2}$  & $\stdx{1.1}$  & $\stdx{0.8}$  & $\stdx{0.0}$  & $\stdx{1.3}$  & $\stdx{1.9}$  & $\stdx{0.4}$ \\ 

\multirow{2}{*}{DeeplabV3+} & \multirow{2}{*}{{\thedataset}}  & mean & $\bbest{78.8}$  & $26.1$  & $\bbest{72.8}$  & $10.4$  & $\bbest{34.2}$  & $\bbest{37.7}$  & $43.3$  & $\bbest{56.3}$  & $\bbest{81.0}$  & $36.1$  & $\bbest{87.8}$  & $\bbest{67.7}$  & $\bbest{46.4}$  & $66.1$  & $\bbest{42.1}$  & $\bbest{33.2}$  & $\bbest{15.3}$  & $\bbest{42.5}$  & $\bbest{52.0}$  & $\bbest{48.9}$ \\ 
& & std & $\stdx{0.6}$  & $\stdx{1.5}$  & $\stdx{1.2}$  & $\stdx{0.9}$  & $\stdx{0.6}$  & $\stdx{0.9}$  & $\stdx{0.7}$  & $\stdx{0.4}$  & $\stdx{0.6}$  & $\stdx{0.9}$  & $\stdx{1.1}$  & $\stdx{0.4}$  & $\stdx{0.6}$  & $\stdx{0.4}$  & $\stdx{0.5}$  & $\stdx{0.7}$  & $\stdx{3.2}$  & $\stdx{0.9}$  & $\stdx{0.3}$  & $\stdx{0.4}$ \\  \hline

\multirow{2}{*}{DeeplabV3+} & \multirow{2}{*}{GTAV+Synscapes} & mean & $72.9$  & $33.3$  & $80.2$  & $24.7$  & $36.5$  & $42.9$  & $57.5$  & $58.0$  & $\bbest{78.3}$  & $41.0$  & $92.6$  & $73.5$  & $39.8$  & $\bbest{87.3}$  & $46.9$  & $43.1$  & $12.1$  & $52.3$  & $48.6$  & $53.8$ \\ 
& & std & $\stdx{2.2}$  & $\stdx{2.1}$  & $\stdx{1.1}$  & $\stdx{4.5}$  & $\stdx{0.7}$  & $\stdx{1.1}$  & $\stdx{0.5}$  & $\stdx{4.3}$  & $\stdx{0.4}$  & $\stdx{2.4}$  & $\stdx{1.2}$  & $\stdx{1.1}$  & $\stdx{0.5}$  & $\stdx{0.9}$  & $\stdx{1.7}$  & $\stdx{2.3}$  & $\stdx{4.7}$  & $\stdx{2.4}$  & $\stdx{1.2}$  & $\stdx{1.4}$ \\ 

\multirow{2}{*}{DeeplabV3+} & \multirow{2}{*}{GTAV+Synscapes+{\thedataset}}  & mean & $\bbest{86.9}$  & $\bbest{50.9}$  & $\bbest{83.1}$  & $\bbest{32.2}$  & $\bbest{44.7}$  & $\bbest{50.4}$  & $\bbest{59.4}$  & $\bbest{66.8}$  & $\bbest{78.5}$  & $\bbest{43.5}$  & $\bbest{94.4}$  & $\bbest{76.1}$  & $\bbest{52.9}$  & $\bbest{87.9}$  & $\bbest{58.0}$  & $\bbest{61.4}$  & $\bbest{18.5}$  & $\bbest{58.4}$  & $\bbest{65.3}$  & $\bbest{61.5}$ \\ 
& & std & $\stdx{0.7}$  & $\stdx{0.8}$  & $\stdx{0.3}$  & $\stdx{0.5}$  & $\stdx{1.0}$  & $\stdx{0.3}$  & $\stdx{0.6}$  & $\stdx{0.7}$  & $\stdx{0.4}$  & $\stdx{0.5}$  & $\stdx{0.3}$  & $\stdx{0.2}$  & $\stdx{1.1}$  & $\stdx{0.5}$  & $\stdx{1.4}$  & $\stdx{1.8}$  & $\stdx{3.0}$  & $\stdx{1.1}$  & $\stdx{0.4}$  & $\stdx{0.5}$ \\ \hline 

\multirow{2}{*}{Co-training} & \multirow{2}{*}{GTAV}  & mean & $\bbest{88.2}$  & $\bbest{44.2}$  & $\bbest{85.3}$  & $\bbest{31.8}$  & $35.5$  & $\bbest{49.3}$  & $\bbest{58.0}$  & $63.1$  & $75.8$  & $51.2$  & $93.2$  & $\bbest{74.4}$  & $50.6$  & $\bbest{88.7}$  & $51.5$  & $31.3$  & $\bbest{11.8}$  & $46.4$  & $45.9$  & $56.6$ \\ 
\UDAbackbone{(DeepLabV3+)} & & std & $\stdx{0.1}$  & $\stdx{0.1}$  & $\stdx{0.1}$  & $\stdx{0.8}$  & $\stdx{0.3}$  & $\stdx{0.3}$  & $\stdx{0.3}$  & $\stdx{0.3}$  & $\stdx{0.2}$  & $\stdx{0.5}$  & $\stdx{0.1}$  & $\stdx{0.2}$  & $\stdx{0.6}$  & $\stdx{0.3}$  & $\stdx{0.2}$  & $\stdx{1.6}$  & $\stdx{2.5}$  & $\stdx{0.6}$  & $\stdx{0.6}$  & $\stdx{0.3}$ \\ 
\multirow{2}{*}{Co-training} & \multirow{2}{*}{Synscapes}  & mean & $78.0$  & $36.3$  & $73.5$  & $16.9$  & $33.8$  & $\bbest{48.4}$  & $56.0$  & $71.0$  & $\gbbest{86.5}$  & $\bbest{56.6}$  & $\gbbest{96.6}$  & $\bbest{74.4}$  & $38.6$  & $\bbest{88.6}$  & $0.6$  & $18.5$  & $0.2$  & $48.5$  & $59.5$  & $51.7$ \\ 
\UDAbackbone{(DeepLabV3+)} & & std & $\stdx{0.2}$  & $\stdx{0.2}$  & $\stdx{0.0}$  & $\stdx{0.0}$  & $\stdx{0.0}$  & $\stdx{0.2}$  & $\stdx{0.9}$  & $\stdx{0.5}$  & $\stdx{0.1}$  & $\stdx{0.0}$  & $\stdx{0.0}$  & $\stdx{0.1}$  & $\stdx{1.6}$  & $\stdx{0.1}$  & $\stdx{0.1}$  & $\stdx{0.1}$  & $\stdx{0.0}$  & $\stdx{0.4}$  & $\stdx{0.0}$  & $\stdx{0.2}$ \\
\multirow{2}{*}{Co-training} & \multirow{2}{*}{{\thedataset}}  & mean & $80.0$  & $36.6$  & $84.1$  & $24.5$  & $\bbest{44.5}$  & $47.5$  & $\bbest{58.6}$  & $\bbest{73.3}$  & $83.4$  & $43.1$  & $\gbbest{96.5}$  & $\bbest{74.0}$  & $\gbbest{59.7}$  & $63.4$  & $\bbest{63.5}$  & $\bbest{69.9}$  & $7.8$  & $\bbest{57.2}$  & $\bbest{63.8}$  & $\bbest{59.5}$ \\  
\UDAbackbone{(DeepLabV3+)} & & std & $\stdx{0.1}$  & $\stdx{0.2}$  & $\stdx{0.1}$  & $\stdx{0.4}$  & $\stdx{0.1}$  & $\stdx{0.1}$  & $\stdx{0.2}$  & $\stdx{0.2}$  & $\stdx{0.1}$  & $\stdx{0.1}$  & $\stdx{0.0}$  & $\stdx{0.5}$  & $\stdx{0.9}$  & $\stdx{0.4}$  & $\stdx{1.9}$  & $\stdx{1.5}$  & $\stdx{1.3}$  & $\stdx{1.3}$  & $\stdx{0.2}$  & $\stdx{0.1}$ \\ \hline

\multirow{2}{*}{Co-training} & \multirow{2}{*}{GTAV+Synscapes+{\thedataset}}   & mean & $\gbbest{91.7}$  & $\gbbest{55.2}$  & $\gbbest{87.2}$  & $\gbbest{42.3}$  & $\gbbest{56.1}$  & $\gbbest{54.1}$  & $\gbbest{66.6}$  & $\gbbest{74.7}$  & $\gbbest{86.2}$  & $\gbbest{57.8}$  & $\gbbest{97.1}$  & $\gbbest{79.1}$  & $\gbbest{59.9}$  & $\gbbest{92.6}$  & $\gbbest{70.1}$  & $\gbbest{75.4}$  & $\gbbest{32.0}$  & $\gbbest{61.5}$  & $\gbbest{68.5}$  & $\gbbest{68.8}$ \\ 
\UDAbackbone{(DeepLabV3+)} & & std & $\stdx{0.1}$  & $\stdx{0.6}$  & $\stdx{0.0}$  & $\stdx{0.6}$  & $\stdx{0.1}$  & $\stdx{0.2}$  & $\stdx{0.4}$  & $\stdx{0.1}$  & $\stdx{0.0}$  & $\stdx{0.2}$  & $\stdx{0.0}$  & $\stdx{0.1}$  & $\stdx{0.4}$  & $\stdx{0.0}$  & $\stdx{0.6}$  & $\stdx{0.8}$  & $\stdx{2.7}$  & $\stdx{1.3}$  & $\stdx{0.9}$  & $\stdx{0.2}$ \\ \hline

\rowcolor{gray!20}  &   & mean & ${96.4}$  & ${79.8}$  & ${91.0}$  & ${57.0}$  & ${68.4}$  & ${65.6}$  & ${71.8}$  & ${82.6}$  & ${92.2}$  & ${73.8}$  & ${98.6}$  & ${80.9}$  & ${64.2}$  & ${93.7}$  & ${77.2}$  & ${81.0}$  & ${56.6}$  & ${64.6}$  & ${71.3}$  & ${77.2}$ \\ 
\rowcolor{gray!20} \multirow{-2}{*}{DeeplabV3+}& \multirow{-2}{*}{Mapillary Vistas (\bbest{reference})} & std & $\stdx{0.2}$  & $\stdx{0.9}$  & $\stdx{0.1}$  & $\stdx{0.8}$  & $\stdx{0.2}$  & $\stdx{0.1}$  & $\stdx{0.1}$  & $\stdx{0.3}$  & $\stdx{0.0}$  & $\stdx{0.6}$  & $\stdx{0.0}$  & $\stdx{0.7}$  & $\stdx{1.8}$  & $\stdx{0.2}$  & $\stdx{0.3}$  & $\stdx{1.2}$  & $\stdx{2.9}$  & $\stdx{1.7}$  & $\stdx{1.0}$  & $\stdx{0.4}$ \\
\noalign{\hrule height 2pt}

\end{tabular}
}

\end{table*}

\subsection{Implementation details}
\label{ssec:implementation}
We perform synth-to-real LAB space alignment as a best practice preprocessing that significantly reduces visual domain gaps \cite{he2021multisource, gomez2023cotraining}. Here, combining different datasets means aggregating them to become a single but multi-source synthetic domain. During the selection of images to compose training mini-batches, we take into account that different sources have different numbers of images to ensure the same probability of selection per image, regardless of the image source.

We use DeepLabV3+ \cite{chen2017DeepLab} and SegFormer MiT-B5 \cite{xie2021segformer} semantic segmentation networks from Detectron2 \cite{wu2019detectron2} and MMSegmentation \cite{mmseg:2020} frameworks, respectively. We use co-training \cite{gomez2023cotraining} and HRDA \cite{hoyer2022hrda} as state-of-the-art methods performing synth-to-real UDA. In \cite{gomez2023cotraining}, co-training results are presented for DeepLabV3+, performing UDA from (GTAV+Synscapes) to Cityscapes, BDD100K, and Mapillary Vistas. In \cite{hoyer2022hrda}, HRDA results are presented for DAFormer \cite{hoyer2022daformer}, derived from SegFormer, and UDA is performed only from GTAV to Cityscapes. 

We consider models trained on synthetic data as baselines. All the baselines involving DeepLabV3+ are trained for 90K iterations with SGD optimizer with a starting learning rate of 0.002 and momentum of 0.9. For Cityscapes, BDD100K, and Mapillary Vistas, we crop the training images to $1024\times512$ pixels, $816\times608$, and $1280\times720$; and batch sizes are set to 8, 16, and 4 images, respectively. SegFormer baselines are trained for 160K iterations with a batch size of 2 in all configurations, using the same hyper-parameters, optimizer, and scheduler reported in \cite{xie2021segformer}, with the same cropping adjustments as for the DeepLabV3+ baselines. For co-training, we use the hyper-parameters reported in \cite{gomez2023cotraining}, except for the batch size that we have been able to increase to the same values used here for the DeepLabV3+ baselines. The crop sizes used in \cite{gomez2023cotraining} and here are the same. For HRDA, we use the settings stated in \cite{hoyer2022hrda}, except for crop sizes, which are set to match our SegFormer baselines. In addition, since our focus is on using multiple synthetic sources, following the instructions of the HRDA framework, we have deactivated the thing-class ImageNet feature distance (FD) inherent to DAFormer-based approaches, which is tailored for the GTAV-to-Cityspaces UDA case.

\subsection{Results analysis}
\label{ssec:resana}

This analysis relies on the results reported in Tables \ref{tab:cityscapes_quantitative}, \ref{tab:bdd_quantitative}, and \ref{tab:mapillary_quantitative}, which correspond to Cityscapes, BDD100K, and Mapillary Vistas, respectively. Each table has a block for approaches based on DeepLabV3+ and another for those based on SegFormer, each including sub-blocks (separated by horizontal lines) for baselines, UDA, and the block reference (training and testing in the same real-world domain). 

Comparing the individual performance of {\thedataset}, GTAV, and Synscapes, we see that the best per-class accuracy is not always given by the same dataset. However, in terms of mIoU, {\thedataset} is the best performing in all cases (always for DeepLabV3+; with Cityscapes and Mapillary Vistas for SegFormer) but one (with BDD100K for SegFormer, in pair with GTAV). The same outcome is observed when applying single-source synth-to-real UDA. In terms of baselines, we see that by adding {\thedataset} to GTAV and Synscapes, the mIoU increases significantly: $6.3$ points for SegFormer/Cityscapes, $8.3$ for DeepLabV3+/Cityscapes, $3.7$ for SegFormer/BDD100K, $2.9$ for DeepLabV3+/BDD100K, $6.0$ for SegFormer/Mapillary Vistas, $7.7$ for DeepLabV3+/Mapillary Vistas. Finally, we see that Musketeers is also beneficial for synth-to-real UDA. For Cityscapes, HRDA reaches mIoU=$76.6$, which is just $4.9$ points behind its reference, while co-training reaches mIoU=$76.7$, just $3.1$ points behind its reference. For BDD100K, HRDA reaches mIoU=$58.6$, $2.9$ points behind its reference, while co-training reaches mIoU=$55.3$, $5.9$ points behind its reference. For Mapillary Vistas, HRDA reaches mIoU=$68.0$, $11.5$ points behind its reference, while co-training reaches mIoU=$68.8$, $8.4$ points behind its reference. 

It is noticeable that Musketeers with HRDA (without ImageNet distance loss) reach mIoU=$76.6$ for Cityscapes, which is the new synth-to-real state-of-the-art to the best of our knowledge. The mIoU=$75.2$ reached by co-training is $5.0$ points better than the results reported in \cite{gomez2023cotraining}. For the case of BDD100K and Mapillary Vistas, co-training improves $5.2$ and $7.5$ points from the results reported in \cite{gomez2023cotraining}. 

\begin{figure*}
    \centering
    \includegraphics[width=0.94\textwidth]{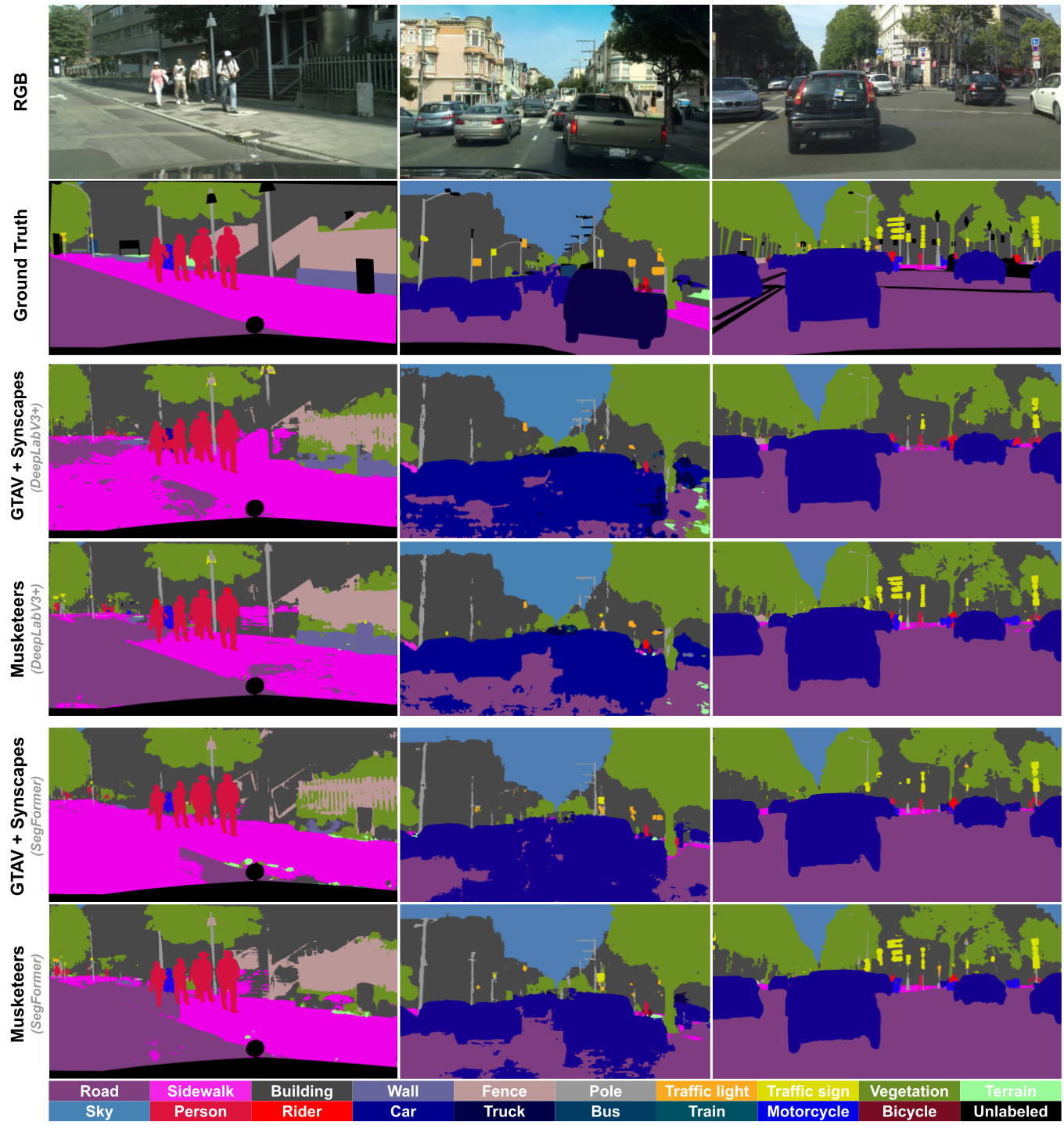}
    \caption{From left to right, each column considers Cityscapes, BDD100K, and Mapillary Vistas. We show three blocks of two rows. First: RGB images and semantic ground truth. Second: semantic segmentation inference using DeepLabV3+, training with GTA+Synscapes and with the Musketeers (+{\thedataset}). Third: as the second block but using SegFormer.}
    \label{fig:qualitative_citys}
    \vspace{-4mm}
\end{figure*}

\Fig{qualitative_citys} illustrates how {\thedataset} is helping to provide stronger baselines as part of the Musketeers, no matter if we use DeepLabV3+ or SegFormer. For Cityscapes, {\thedataset} helps to better separate the road from the sidewalk; for BDD100K, it helps to segment the vehicles better; and, for Mapillary Vistas, traffic signs are better-segmented thanks to {\thedataset}. In addition, looking at the qualitative results of the different validation sets, what we observe is that by using {\thedataset}, we improve the segmentation of objects that occupy a small image area ({\eg}, traffic signs and lights, poles, pedestrians, {\etc}), which is usually the case when these objects are at mid and large distances from the ego-vehicle. 

Overall, we think that {\thedataset}, individually or as part of Musketeers, can become a core contribution to developing new synth-to-real UDA proposals for onboard vision.    

\begin{figure}
    \centering
    \includegraphics[width=0.96\columnwidth]{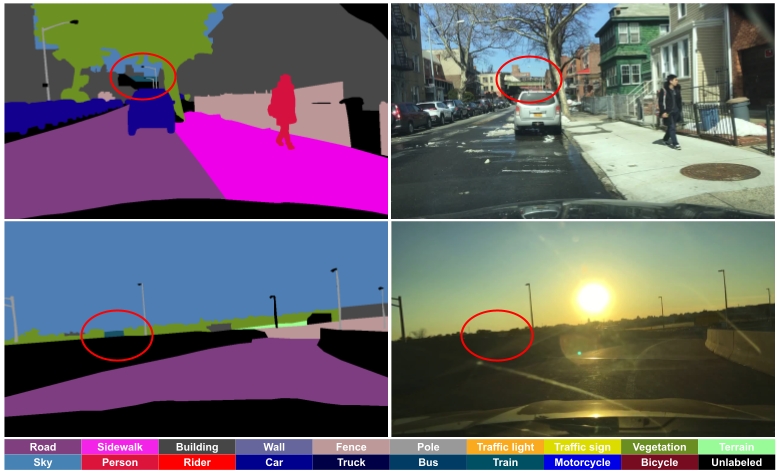}
    \caption{BDD100K images from the validation set containing very challenging Train class samples (encircled in red).}
    \label{fig:bdd_train_class_challenging_samples_challenge}
\end{figure}
\begin{figure}
    \centering
    \includegraphics[width=0.96\columnwidth]{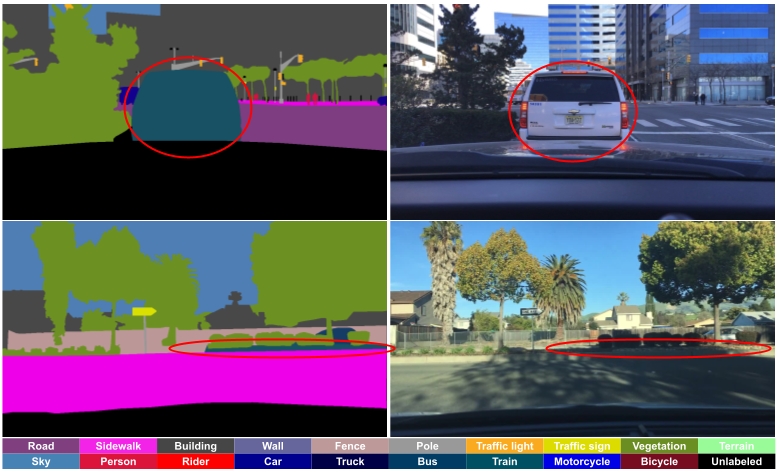}
    \caption{BDD100K images from the validation set containing wrong labels of the Train class (encircled in red).}
    \label{fig:bdd_train_class_wrong_labels}
\end{figure}

\begin{figure}
    \centering
    \includegraphics[width=0.96\columnwidth]{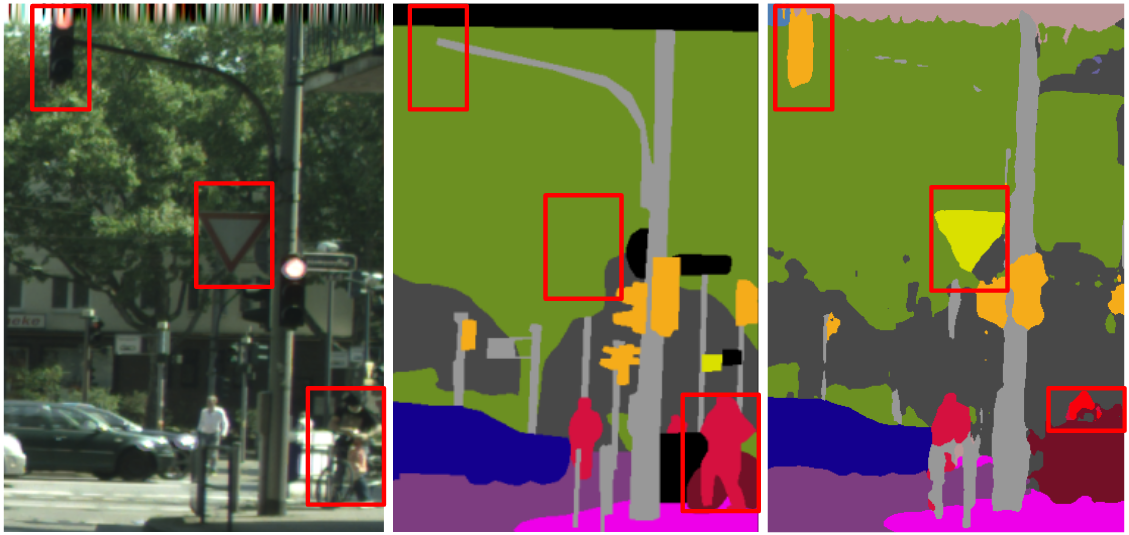}
    \caption{Left to right: RGB image from Cityscapes' validation set, its semantic ground truth, and its co-training inference. Human labeling misses Traffic Light and Traffic Sign instances that co-training labels (boxed in red). In addition, a rider in the bottom right of the image is incorrectly labeled as a Pedestrian by humans, while co-training identifies it as Rider + Bicycle.}
    \label{fig:wrong_labels_cityscapes_01}
\end{figure}

\subsection{Erroneous human labels}
\label{ssec:errorshumanlabels}

\begin{figure}
    \centering
    \includegraphics[width=\columnwidth]{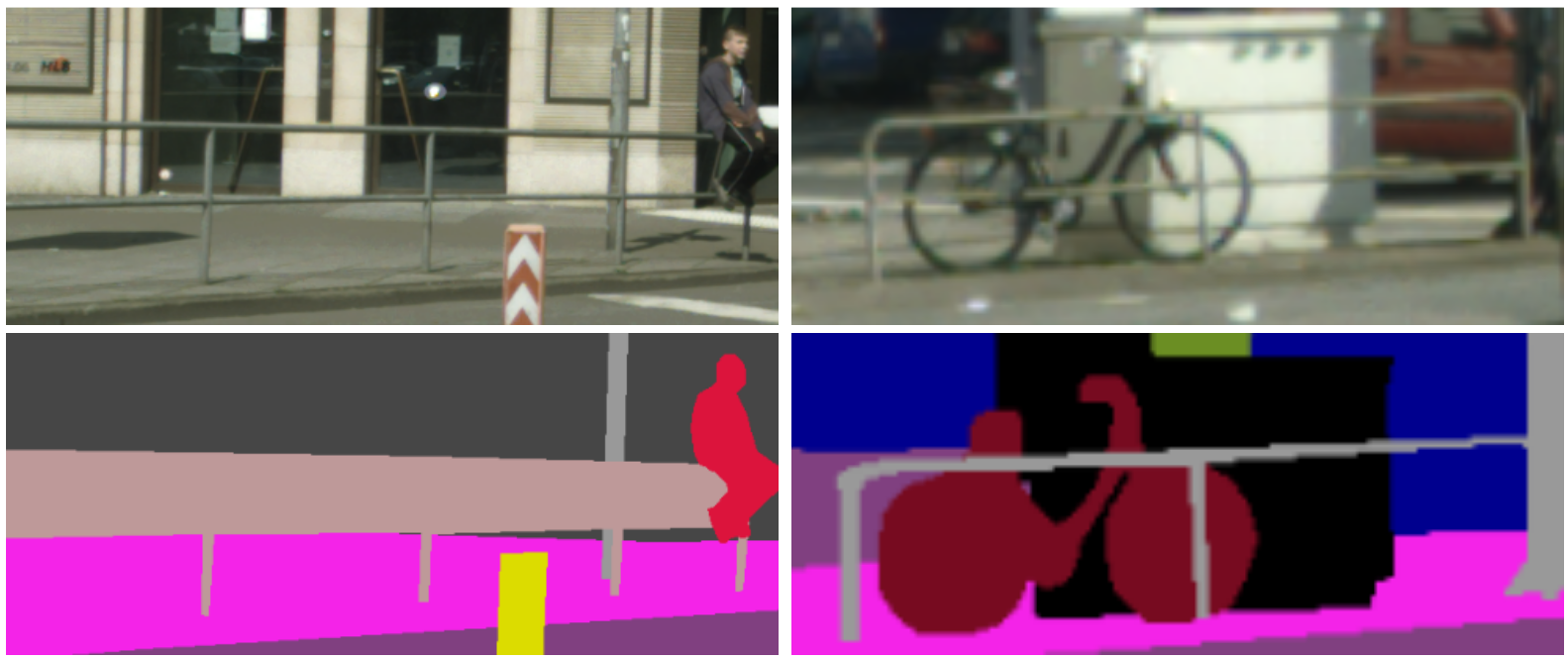}
    \caption{Cityscapes images from the validation set containing inconsistencies in human labeling for the Fence class. Note how different criteria have been used to label the Fences in the left and right images.}
    \label{fig:wrong_labels_cityscapes_02}
\end{figure}

\begin{figure}
    \centering
    \includegraphics[width=\columnwidth]{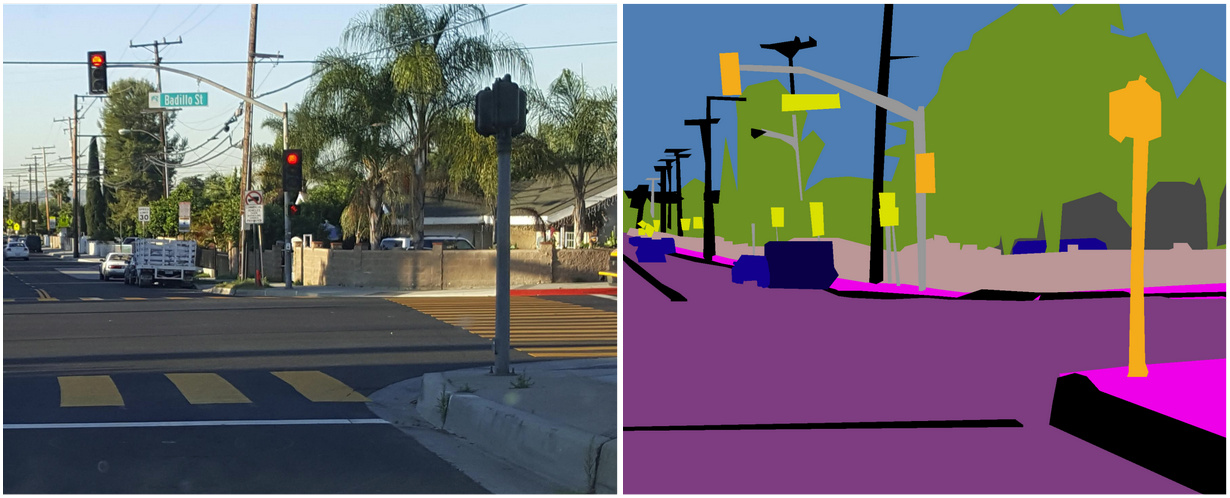}
    \caption{The closest Traffic Light instance is fully labeled as such (in yellow the lightbox and its support), while for the further away only the light boxes are labeled as Traffic Light.}
    \label{fig:wrong_labels_mapillary}
\end{figure}

During the qualitative analysis of the results, we observed labeling errors in the ground truth of the real-world datasets. For instance, the BDD100K validation set only contains seven images (out of 1,000 images) with labels of the Train class. In five of these, the samples are extremely challenging (\Fig{bdd_train_class_challenging_samples_challenge}), while in the other two, the labels are just wrong (\Fig{bdd_train_class_wrong_labels}). Analyzing the BDD100K training set, we found only 27 images with labels of the Train class (out of 7,000 images), where the labels were wrong in 13 of them. All this explains the poor results seen in Table \ref{tab:bdd_quantitative} for the Train class. As can be seen in Figures \ref{fig:wrong_labels_cityscapes_01} and \ref{fig:wrong_labels_cityscapes_02}, Cityscapes also have some errors and inconsistencies regarding human labeling. In \Fig{wrong_labels_cityscapes_01}, we see how the co-training method properly labels a traffic light and a traffic sign, while humans did not label them. Since this image is from the validation set of Cityscapes, the self-labels are considered wrong. In the same image, a rider on a bicycle is labeled as a fully visible Person by humans, while the upper part of the person is labeled as a Rider by co-training, and the rest is labeled as a Bicycle. Again, we think that the self-labels better represent this image patch's semantics. In \Fig{wrong_labels_cityscapes_02}, we can see how different criteria have been used to label Fence instances in different images. \ref{fig:wrong_labels_mapillary} shows another example of inconsistency, in this case from a Mapillary Vistas image. We can see how the closed Traffic Light instance has been fully labeled as such (lightbox and its support), while in the further away ones, only the light boxes are labeled as Traffic Light.

\section{Conclusions}
\label{sec:conclusions}

This paper introduces {\thedataset}, a diverse, compact, and photorealistic dataset of synthetic images capturing urban driving scenes. It comes with pixel-wise ground truth for depth estimation and semantic and instance segmentation. A 2D BB and a degree of occlusion complement instance labels. {\thedataset} brings differences from GTAV and Synscapes regarding road layouts, background content, and traffic participants. All of them form the Three Musketeers dataset. We have demonstrated the value of {\thedataset} individually and as part of the Musketeers by establishing new benchmarks on synth-to-real UDA for onboard semantic segmentation, where we have used state-of-the-art methods with core differences (HRDA, co-training) and a variety of real-world target domains (Cityscapes, Mapillary Vistas, BDD100K). We make {\thedataset} openly and freely accessible to the research community. Therefore, we expect that the research community can use {\thedataset} not only for tasks related to semantic segmentation but also for instance segmentation, object detection, and monocular depth estimation. As future work, we plan to work on synth-to-real UDA for the instance segmentation task, where pixels must not only be assigned a semantic class but also an individual identifier (\eg, to distinguish one vehicle from another). For this task, it is essential that {\thedataset} not only provides class labels but instance labels too. In this regard, {\thedataset} offers fine-grained instance information as distinguishing instances of motorbikes/bikes from their human riders in contrast to Synscapes' instances (where the rider and the motorbike/bike form the same instance) and GTAV-based instances (no bikes/cyclists are included, and motorbike riders are treated as pedestrians). In addition, as future work, we want to explore how {\thedataset} can help in synth-to-real active learning procedures, where, in contrast to UDA, this approach allows for a certain budget to label a few real-world training images by hand. The challenge lies in selecting which images to allocate that budget.

\section*{Accessing the dataset}
In order to download {\thedataset}, users have two options, either accessing the {\thedataset} web page, \url{https://urbansyn.org}, or accessing the popular HuggingFace services, \url{https://huggingface.co/datasets/UrbanSyn/UrbanSyn}. Both sites include all the information required to download the data, a description of their content, and visual examples.

\section*{Acknowledgements}
This work has been supported by the Spanish grants Ref. PID2020-115734RB-C21 (ADA/SSL-ADA subproject) and PID2020-115734RB-C22 (ADA/PGAS-ADA subproject), both funded by MCIN/AEI/10.13039/501100011033. 
Antonio M. López acknowledges the financial support to his general research activities given by ICREA under the ICREA Academia Program. CVC's authors acknowledge the support of the Generalitat de Catalunya CERCA Program and its ACCIO agency to CVC’s general activities. Jose A. Iglesias-Guitian acknowledges the financial support to his general research activities given by UDC-Inditex InTalent programme, the Spanish Ministry of Science and Innovation (AEI/RYC2018-025385-I), and Xunta de Galicia (ED431F 2021/11, EU-FEDER ED431G 2019/01).

\newpage
\bibliographystyle{IEEEtran}
\bibliography{references}

\end{document}